\documentclass[a4paper,fleqn]{cas-sc}

\usepackage[authoryear,longnamesfirst]{natbib}
\usepackage{natbib}
\setcitestyle{numbers,square}

\def\tsc#1{\csdef{#1}{\textsc{\lowercase{#1}}\xspace}}
\tsc{WGM}
\tsc{QE}

\begin{document}
\let\WriteBookmarks\relax
\def\floatpagepagefraction{1}
\def\textpagefraction{.001}

% Short title
\shorttitle{A Comprehensive Overview from Training to Inference}    

% Short author
\shortauthors{Yiheng Liu et al.}  

% Main title of the paper
\title [mode = title]{Understanding LLMs: A Comprehensive Overview from Training to Inference}  

\author[1]{Yiheng Liu}
\author[1]{Hao He}
\author[1]{Tianle Han}
\author[1]{Xu Zhang}
\author[1]{Mengyuan Liu}
\author[1]{Jiaming Tian}
\author[2]{Yutong Zhang}
\author[3]{Jiaqi Wang}
\author[4]{Xiaohui Gao}
\author[4]{Tianyang Zhong}
\author[5]{Yi Pan}
\author[5]{Shaochen Xu}
\author[5]{Zihao Wu}
\author[5]{Zhengliang Liu}
\author[2]{Xin Zhang}
\author[3]{Shu Zhang}
\author[4]{Xintao Hu}
\author[4]{Tuo Zhang}

\author[1]{Ning Qiang}

\author[5]{Tianming Liu}
\author[1]{Bao Ge \corref{cor1}}

\cortext[cor1]{Corresponding author}

\affiliation[1]{organization={School of Physics and Information Technology, Shaanxi Normal University},
        city={Xi'an},
        postcode={710119}, 
        state={Shaanxi},
        country={China}}

\affiliation[2]{organization={Institute of Medical Research, Northwestern Polytechnical University},
        city={Xi'an},
        postcode={710072},
        state={Shaanxi},
        country={China}}

\affiliation[3]{organization={School of Computer Science, Northwestern Polytechnical University},
        city={Xi'an},
        postcode={710072},
        state={Shaanxi},
        country={China}}
        
\affiliation[4]{organization={School of Automation, Northwestern Polytechnical University},
        city={Xi'an},
        postcode={710072},
        state={Shaanxi},
        country={China}}

\affiliation[5]{organization={School of Computing, The University of Georgia},
city={Athens},
postcode={30602}, 
country={USA}}

% Here goes the abstract
\begin{abstract}
    The introduction of ChatGPT has led to a significant increase in the utilization of Large Language Models (LLMs) for addressing downstream tasks. There's an increasing focus on cost-efficient training and deployment within this context. Low-cost training and deployment of LLMs represent the future development trend. This paper reviews the evolution of large language model training techniques and inference deployment technologies aligned with this emerging trend. The discussion on training includes various aspects, including data preprocessing, training architecture, pre-training tasks, parallel training, and relevant content related to model fine-tuning. On the inference side, the paper covers topics such as model compression, parallel computation, memory scheduling, and structural optimization. It also explores LLMs' utilization and provides insights into their future development.
\end{abstract}

% Use if graphical abstract is present
%\begin{graphicalabstract}
%\includegraphics{}
%\end{graphicalabstract}

% Research highlights
% \begin{highlights}

% \end{highlights}

% Keywords
% Each keyword is seperated by \sep

\maketitle

\section{Introduction}
Language modeling (LM) is a fundamental approach for achieving cognitive intelligence in the field of natural language processing (NLP), and its progress has been notable in recent years \cite{liu2023summary,wang2023prompt,zhao2023survey}. It assumes a central role in understanding, generating, and manipulating human language, serving as the cornerstone for a diverse range of NLP applications \cite{kaddour2023challenges}, including machine translation, chatbots, sentiment analysis, and text summarization. With the evolution of deep learning, the early statistical language models (SLM) have gradually transformed into neural language models (NLM) based on neural networks. This shift is characterized by the adoption of word embeddings, representing words as distributed vectors. Notably, these word embeddings have consistently excelled in practical NLP tasks, profoundly shaping the field's progress. Pre-trained language models (PLM) represent a subsequent phase in the evolution of language models following NLM. Early attempts at PLMs included ELMo \cite{ELMo}, which was built on a Bidirectional LSTM architecture. However, with the advent of the transformer architecture \cite{transformer}, characterized by parallel self-attention mechanisms, the pre-training and fine-tuning learning paradigm has propelled PLM to prominence as the prevailing approach. These models are typically trained via self-supervision on extensive datasets, cementing their status as the primary methodology in the field. 

The Transformer architecture is exceptionally well-suited for scaling up models, and research analysis has revealed that increasing the model's scale or training data size can significantly enhance its performance. Many studies have pushed the boundaries of model performance by continuously expanding the scale of PLM \cite{GPT-2,GPT3,touvron2023llama,touvron2023llama2}. As models grow larger, a remarkable phenomenon known as "emergence" occurs, wherein they exhibit astonishing performance \cite{GPT3}. These models are capable of generating high-quality text and possess robust learning and reasoning abilities. They can even tackle few-shot learning tasks through in-context learning (ICL) \cite{GPT3}. This remarkable capability enables their seamless application to a wide range of downstream tasks across diverse domains \cite{10.1007/978-3-031-21014-3_28,PMID:36097765,liao2023maskguided,rezayi2023exploring}. 

Pre-trained language models (PLMs) with significantly larger parameter sizes and extensive training data are typically denoted as Large Language Models (LLMs)  \cite{zhong2023chatradiovaluer,liu2023evaluating,zhong2023chatabl}. The model size usually exceeds 6-10 billion (6-10B) parameters. A prominent milestone in the development of LLMs is exemplified by the GPT series \cite{GPT,GPT-2,GPT3,openai2023gpt4}. Notably, OpenAI released ChatGPT in November 2022, marking a pivotal moment in the era of LLMs and a game-changing moment in the field of artificial intelligence. ChatGPT has empowered current AI algorithms to achieve unprecedented levels of strength and effectiveness, reshaping the way humans employ or develop AI algorithms. Its emergence has captured the attention of the research community. However, owing to ChatGPT's absence as an open-source platform, the principal way to use ChatGPT currently is by accessing it through OpenAI's website at \url{https://chat.openai.com} or via their API interface. Training LLMs that can serve as alternatives to ChatGPT, or domain-specific LLMs, has become highly necessary \cite{dai2023auggpt,liu2023deidgpt,ma2023impressiongpt,liao2023differentiate,dai2023adautogpt,liu2023summary,guan2023cohortgpt,liu2023pharmacygpt}. Training and deploying LLMs demand expertise in handling large-scale data and substantial practical experience in distributed parallel training  \cite{wei2023chat2brain,10.1007/978-3-031-43907-0_40,10.1093/psyrad/kkad011}. This requirement emphasizes the need for researchers developing LLMs to possess significant engineering capabilities in addressing the challenges encountered during LLM development. Researchers who are interested in the field of LLMs must possess engineering skills or learn to collaborate effectively with engineers.

For the above reasons, the primary objective of this paper is to provide a comprehensive overview of LLMs training and inference techniques to equip researchers with the knowledge required for developing, deploying, and applying LLMs. The structure of the rest of this review is as follows: In Section~\ref{sec::background}, we will introduce the relevant background and foundational knowledge of LLMs. In Section~\ref{sec::training}, we will delve into the technical aspects of training LLMs, while in Section~\ref{sec::inference} we will explore the technologies related to LLM's inference and deployment. In Section~\ref{sec::app}, we will discuss the utilization of LLMs, and Section~\ref{sec::future} will explore the future directions and their implications for LLMs.

\section{Background Knowledge}
\label{sec::background}
\subsection{Transformer}
Transformer is a deep learning model based on an attention mechanism for processing sequence data that can effectively solve complex natural language processing problems. This model was first proposed in 2017 \cite{transformer}, and replaced the traditional recurrent neural network architecture \cite{sutskever2014sequence} in machine translation tasks as the state-of-the-art model at that time. Due to its suitability for parallel computing and the complexity of the model itself, Transformer outperforms the previously popular recurrent neural networks in terms of accuracy and performance. The Transformer architecture consists primarily of two modules, an Encoder and a Decoder, as well as the attention mechanism within these modules.

\subsubsection{Self-Attention}
\textbf{Self-Attention Structure \cite{transformer}:} Essentially, the attention mechanism aims at selecting a small amount of important information from a large amount of data and focusing on these important pieces while ignoring the majority of unimportant information. The self-attention mechanism, as a variant of the attention mechanism, reduces reliance on external information and excels at capturing internal correlations within data or features. Applying the self-attention mechanism in text-primarily involves calculating the mutual influence between words to address the issue of long-range dependencies. Additionally, self-attention is the core idea behind transformers. The core formula for key-value attention is as follows:
\begin{equation}
    Attention(Q, K, V) = softmax(\frac{QK^{T}}{\sqrt{d_{k}}})V
\end{equation}
Self-attention allows the model to weigh the importance of different words in a sentence when predicting a particular word. It calculates a weighted sum of the values of all words in the sentence, where the weights are determined by the relevance of each word to the target word.

The self-attention mechanism consists of three steps: calculating the query, key, and value vectors. The query vector represents the word being attended to, while the key vectors represent all the words in the sentence. The value vectors store the information associated with each word. The attention weights are computed by taking the dot product between the query and key vectors, followed by a softmax operation to obtain a distribution over the words.

\textbf{Multi-Head Attention \cite{transformer}:} Multi-head self-attention extends the self-attention mechanism by performing it multiple times in parallel. Each attention head learns to focus on different aspects of the input, capturing different dependencies and patterns. The outputs of the attention heads are then concatenated and linearly transformed to obtain the final representation. By using multiple attention heads, the model can capture both local and global dependencies, allowing for a more comprehensive understanding of the input sequence. This parallelization also enhances the model's capacity to capture complex relationships between words. The Multi-head attention can be formulated as follows:

\begin{equation}
    \begin{aligned}
        MultiHeadAttention(Q, K, V) = Concat[head_{1}, \ldots, head_{h}]W^{o} \\
        where \ head_{i}= Attention(QW^{Q}_{i},KW^{K}_{i},VW^{V}_{i})
    \end{aligned}
\end{equation}

In this case, "$Concat$" means to concatenate the attention calculation results of each head, "$W^{o}$" is the weight matrix of the output layer, used to linearly transform the concatenated results. This yields the output of multi-head attention. In summary, multi-head attention enhances the model's ability to represent input sequences by performing parallel attention calculations under different linear transformations, then concatenating and linearly transforming the results. This mechanism plays an important role in the Transformer model, helping to handle long-range dependencies and improve model performance.

\subsubsection{Encoder}

The encoder module \cite{transformer} of the Transformer model is composed of multiple identical layers, each of which includes a multi-head attention mechanism and feed-forward neural network \cite{bebis1994feed}. In the multi-head attention mechanism, each position in the input sequence is calculated for attention with other positions to capture the dependencies between different positions in the input sequence. The feed-forward neural network is then used to further process and extract features from the output of the attention mechanism. The encoder module gradually extracts features of the input sequence through the stacking of multiple such layers and passes the final encoding result to the decoder module for decoding. The design of the encoder module enables it to effectively handle long-range dependencies within the input sequence and has significantly improved performance in various NLP tasks. 

\subsubsection{Decoder}
The decoder module \cite{yang2020sub} of the Transformer model is also composed of multiple identical layers, each of which includes a multi-head attention mechanism and a feed-forward neural network. Unlike the encoder, the decoder also includes an additional encoder-decoder attention mechanism, used to compute attention on the input sequence during the decoding process. At each position, the decoder can only perform self-attention calculations with the positions before it to ensure that the generation of the sequence does not violate grammar rules. Masks play an important role in the decoder, ensuring that only information before the current time step is focused on when generating the output sequence, and not leaking information from future time steps. Specifically, the decoder's self-attention mechanism uses masks to prevent the model from accessing future information when generating predictions at each time step, maintaining the causality of the model. This ensures that the output generated by the model depends on the information at the current time step and before, without being influenced by future information.

\subsubsection{Positional Embedding} 
Position and order are crucial for certain tasks, such as understanding a sentence or a video. Position and order define the grammar of a sentence, they are integral to the semantics of sentences. The Transformer utilizes Multi-Head Self-Attention (MHSA) to avoid the recursive approach of RNN, thus speeding up the training process. Additionally, it can capture long-range dependencies in sentences and handle longer inputs. When each token in a sentence passes through the Transformer's Encoder/Decoder stack, the model itself lacks any sense of position/order for each token (permutation invariance). Therefore, a method is still needed to incorporate the sequential information of tokens into the model. To enable the model to perceive the input sequence, positional information about the location of each token in the sentence can be added, and this technique is known as positional embedding (PE). which is used in the Transformer model to incorporate the sequential order of tokens into the input representation. Since the Transformer does not have recurrent connections, it lacks the inherent notion of token order present in recurrent neural networks. To address this, positional embedding assigns a unique vector to each token position in the input sequence. These positional embeddings are added to the word embedding before being fed into the model. By including positional information, the model can differentiate between tokens based on their position in the sequence. In the Transformer model, the core formula of the position embedding can be expressed as:
\begin{equation}
    PE(pos,2i)=sin(\frac{pos}{10000^{(\frac{2i}{d_{model}})}})
\end{equation}
\begin{equation}
    PE(pos,2i+1)=cos(\frac{pos}{10000^{(\frac{2i}{d_{model}})}})
\end{equation}
In this equation, $PE$ represents the position embedding matrix, $pos$ represents the position of a token in the sentence, $i$ represents the dimension index of the position embedding, and $d_{model}$ represents the hidden layer dimension of the Transformer model. By using sine and cosine functions and performing different calculations on the position (pos) and dimension (i), this formula generates unique position embedding values for each position and dimension. As a result, each token is assigned a unique position embedding vector, allowing the model to perceive the sequential information of tokens in the sentence. In practical applications, the position embedding matrix is added to the input word embedding matrix to combine position information and semantic information, thereby providing a more comprehensive input representation for the Transformer model.

Two commonly used positional encoding methods in Transformer are Absolute Positional Encoding and Relative Positional Encoding.

(1) Absolute Positional Encoding: It generates unique positional embedding values for each position and dimension by using sine and cosine functions. This method uses sine and cosine functions in the mentioned formula to calculate the positional embedding values and adds them to the word embeddings. Absolute Positional Encoding provides a unique encoding for each position, enabling the model to perceive the sequential information of words in the sentence. 

(2) Relative Positional Encoding: It is an encoding method based on relative positional relationships. Relative Positional Encoding represents positional information by calculating the relative distances between words. This method is used in models like Transformer-XL \cite{dai2019xl}, and Relative Positional Encoding can better capture the relative positional relationships between words when dealing with long sequences. Both of these positional encoding methods aim to provide the positional information of words in the input sequence to the Transformer model, enabling the model to better comprehend and process sequential data. The specific choice of positional encoding method depends on the specific application scenario and model design. 

There are also other positional encoding methods applied to other models, such as RoPE \cite{su2023roformer} and ALiBi \cite{press2021train}. 

RoPE is a method that uses Absolute Positional Encoding to represent Relative Positional Encoding and is applied in the design of large language models like PaLM \cite{chowdhery2022palm}, LLaMA \cite{touvron2023llama}, and GLM-130B \cite{zeng2022glm}. 

ALiBi does not add positional embeddings to word embeddings but instead adds a pre-defined bias matrix to the attention score based on the distance between tokens. It is applied in the design of large language models like BLOOM \cite{workshop2022bloom}.

Some other positional encoding methods, such as mixed positional encoding, multi-digit positional encoding, and implicit positional encoding, are also used by some models.

\subsection{Prompt Learning} 
Prompt learning serves as a widely adopted machine learning approach, particularly in the field of NLP. At its core, this methodology involves guiding a model to produce specific behaviors or outputs through the careful design of prompt statements. It is commonly employed to fine-tune and guide pre-trained LLMs for executing particular tasks or generating desired results. Researchers have observed that the design of specific prompt statements can steer pre-trained models to perform various tasks, such as question-answering, text generation, and semantic understanding \cite{ZHAO2023100005,Holmes_2023,wu2023exploring,rezayi2022agribert,liu2023radiologygpt,Liu_2023,wang2023review,li2023artificial,cai2023coarsetofine,dai2023samaug,zhang2023segment,xiao2023instructionvit}. The strength of this approach lies in its ability to adapt to different tasks through simple modifications to prompt statements, eliminating the need for retraining the entire model. For LLMs like the GPT series and other pre-trained models, prompt learning provides a straightforward and powerful means for model fine-tuning. By supplying appropriate prompts, researchers and practitioners can customize the model's behavior, making it more suitable for specific domains or task requirements. In short, prompt learning is a machine learning approach that, builds upon pre-trained language models, and guides the model to perform various tasks through the design of prompt statements, offering increased flexibility for customizing model applications. In this Section, we will introduce the basic knowledge of prompt learning.

 \subsubsection{Background and Overview}
% \begin{equation}
    
% \end{equation}

% Figure~\ref{fig:enter-label}
% \begin{figure}
%     \centering
%     \includegraphics{figures/xxx.}
%     \caption{Captiondsadasd}
%     \label{fig:enter-label}
% \end{figure}
Prompt learning is a new approach to machine learning \cite{liu2023pre}. In the early field of natural language processing (NLP), researchers mainly used fully supervised learning mode\cite{kotsiantis2007supervised}, which trained models for specific tasks on the input and output example dataset of the target task. However, due to the limited training dataset, this method cannot train high-quality models well, so early NLP relied more on feature engineering; With the emergence of neural network models and their use in the field of NLP, people have begun to pay attention to architecture engineering \cite{bengio2013representation}.

However, between 2017 and 2019, the learning approach of NLP models shifted from fully supervised learning to a new mode: pre-train and fine-tune paradigm\cite{dong2019unified}. In this paradigm, a model with a fixed architecture is pre-trained as a language model to predict the probability of observed text data. 
Due to the abundant raw text data required for training language models, these language models can be trained on large datasets. During this process, language models can learn robust universal features of the language they are modeling. Then, by introducing additional parameters and fine-tuning them using task-specific objective functions, the PLM mentioned above will adapt to different downstream tasks. At this point, the focus of research shifted to objective engineering, which is to design training objectives during pre-training and fine-tuning. Since BERT, NLP has been using pre-training and fine-tuning methods for a long period of time, but this approach requires a new model to be fine-tuned for each task and cannot be shared. But for an LLM, it feels like customizing each task, which is very inefficient \cite{liu2023pre}. 

Prompt learning, this method has demonstrated amazing capabilities in GPT-3. The GPT-3 model can handle many tasks with only a few samples by using natural language prompts and task demonstrations as context, without updating parameters in the underlying model. Prompt Learning replaces the process of pre-trained and fine-tuning with pre-trained, prompts and predictions. In this paradigm, the downstream task is not to adapt the pre-trained LM to the downstream task through objective engineering, but to redefine the downstream task with the help of text prompts, making it look more like the tasks solved during the original LM training. For prompt learning, it is only necessary to insert different prompt parameters to adapt to different tasks. That is to say, each task only needs to train the prompt parameter separately, without the need to train the entire pre-trained language model\cite{schick2020s}. This approach greatly improves the efficiency of using pre-trained language models and significantly shortens training time.

\subsubsection{Basic components and process of Prompt learning}  
In the traditional pre-trained+fine-tuning paradigm, there is a gap between the pre-trained stage and downstream tasks \cite{liu2023pre}, while prompt learning can maintain consistency between the pre-trained target format and downstream task output format, that is, align the form of downstream tasks with the form of PLMs pre-trained tasks. When training PLMs, we can transform the original target task into a fill-in-the-blank or continuation task similar to the pre-trained task of PLMs by constructing a prompt. The advantage of this method is that through a series of appropriate prompts, we can use a single language model to solve various downstream tasks.

Prompt learning optimizes the performance of models on different tasks by using pre-trained models and designing appropriate templates. Prompt learning consists of prompt templates, answer mappings, and pre-trained language models. The prompt template is the main body of the prompt, and fill in the blank \cite{petroni2019language} and generate based on prefix \cite{lester2021power}are two common types of prompt learning templates. The fill-in-the-blank template selects one or more positions in the text and represents them with [MASK] tags, used to prompt the model to fill in the corresponding words; Prefix-based template generation involves adding a specific prefix before a sentence to guide the model in generating appropriate text. Answer mapping is the process of evaluating all possible answers according to a probability distribution, selecting the most likely answer as the predicted output, and converting it into appropriate category mapping words. This process typically involves converting labels into natural language vocabulary, known as Verbalizer \cite{schick2020exploiting}.

The workflow of Prompt learning mainly includes the following four parts:

(1)Use PLMs as base encoders

(2)Add additional context (template) with a [MASK] position

(3)Project labels to label words (verbalizer)

(4)Be the GAP between pre-training and fine-tuning

After defining the template and answer space, we need to choose a suitable pre-trained language model. There are now various pre-trained models (PTMs) with good performance, and when selecting a model, one usually considers its paradigm, such as Auto recursive, Masked Language Modeling, Encoder Decoder, etc. Based on this, for the summary task, a more suitable Bidirectional and Auto-Regressive Transformers (BART) model can be selected.

The selection of a template plays a very important role in the prompt learning. Templates can generally be distinguished based on whether they are manually specified: artificially constructed templates or automatically searched templates. Artificially created templates are the most intuitive method, easy to understand, and have good performance in practical applications. However, artificially constructed templates also have some drawbacks: prior knowledge is required when designing templates manually \cite{shin2021constrained}, and there may be failures \cite{jiang2020can}. There are two types of automatically generated templates: discrete prompts and continuous prompts. Discrete prompts allow the model to select the optimal template in a set of discrete template spaces, while continuous prompts allow the language model to automatically train a prompt. According to research, using multiple templates \cite{duh2011generalized} can improve the performance of the model. The simplest way to choose to use multiple templates and aggregate them together to complete an answer is to take the average \cite{jiang2020can} or weighted average of each template output \cite{schick2020exploiting}.

Verbalizer is the process of mapping labels to label words, and the selection of verbalizers is also crucial for prompt learning. There are two ways to construct a verbalizer: manual definition and automatic search. The manual definition requires professional knowledge and may have disadvantages such as strong subjectivity and a small coverage area. To solve this problem, we can choose the following solutions: (1) Manually design with human prior knowledge; (2) Start with an Intel label word, paraphrase and expand; (3) Start with an internal label word, using external knowledge and expand; (4) Decompose the label with multiple tokens; (5) Virtual token and optimize the label embedding. In addition, we can use external knowledge bases to expand and improve label words, thereby achieving better text classification results\cite{jiang2021can}.

\subsubsection{learning strategy}
The emergence of the new paradigm of Prompt learning has brought significant changes to the training process.
The learning strategies for Prompt learning mainly include the following:
(1) Pre-training then fine-tuning, which is a traditional pre-training+fine tuning method \cite{mccloskey1989catastrophic}; (2) Tuning free promotion, relying on the designer LM of prompts to directly provide answers \cite{brown2020language}; (3) Fixed LM prompt tuning, which updates the relevant parameters of prompts using downstream task training data; (4) Fix prompt LM tuning, this strategy is to fine-tune the parameters of LM, which have fixed parameters when using prompts; (5) Prompt+LM tuning is a strategy that updates both prompts related parameters and LM parameters.

These different learning strategies can be selected based on specific tasks and needs. Pre-training + fine-tuning is the most common strategy, suitable for most tasks \cite{mccloskey1989catastrophic}. No fine-tuning prompts are suitable for simple tasks, which can greatly reduce training time and computational resource consumption. Fixed LM prompt fine-tuning and fixed prompt LM fine-tuning are suitable for tasks that require more precise control and can optimize model performance by adjusting prompt parameters or language model parameters. Combining prompts and LM fine-tuning combines the advantages of both and can further improve model performance \cite{liu2023pre}.

In summary, Prompt learning provides us with a new training paradigm that can optimize model performance on various downstream tasks through appropriate prompt design and learning strategies. Choosing the appropriate template, constructing an effective verbalizer, and adopting appropriate learning strategies are all important factors in improving the effectiveness of prompt learning.

\section{Training of Large Language Models}
\label{sec::training}
The training of LLMs can be broadly divided into three steps. The first step involves data collection and processing. The second step encompasses the pre-training process, which includes determining the model's architecture and pre-training tasks and utilizing suitable parallel training algorithms to complete the training. The third step involves fine-tuning and alignment. In this section, we will provide an overview of the model training techniques. This will include an introduction to the relevant training datasets, data preparation and preprocessing, model architecture, specific training methodologies, model evaluation, and commonly used training frameworks for LLMs.

\subsection{Data Preparation and Preprocessing}
\subsubsection{Dataset}

Training LLMs require vast amounts of text data, and the quality of this data significantly impacts LLM performance. Pre-training on large-scale corpora provides LLMs with a fundamental understanding of language and some generative capability. The first step in LLM training is collecting substantial corpora of natural language text. Pre-training data sources are diverse, commonly incorporating web text, conversational data, and books as general pre-training corpora. Additionally, some research efforts introduce specialized data from professional domains, such as code or scientific data, to enhance LLM capabilities in those fields. Leveraging diverse sources of text data for LLM training can significantly enhance the model's generalization capabilities. In the following section, we will present the commonly used datasets for training LLMs as shown in Table~\ref{tab:data}. These corpora are categorized into 5 groups for discussion.

\begin{table}[h]
	\caption{Commonly used corpora information.}
    \label{tab:data}
	\centering
	\begin{tabular}{ccc}
	\toprule
	Corpora & Type & Links \\
	\midrule	
	BookCorpus \cite{BookCorpus} & Books & \url{https://github.com/soskek/bookcorpus}\\	
	Gutenberg \cite{Gutenberg} & Books & \url{https://www.gutenberg.org}   \\	
        Books1 \cite{GPT3} & Books & Not open source yet \\
        Books2 \cite{GPT3} & Books & Not open source yet \\
        CommonCrawl \cite{commoncrawl} & CommonCrawl & \url{https://commoncrawl.org} \\
        C4 \cite{T5} & CommonCrawl & \url{https://www.tensorflow.org/datasets/catalog/c4} \\
        CC-Stories \cite{CC-Stories} & CommonCrawl & Not open source yet\\
        CC-News \cite{liu2019roberta} & CommonCrawl & \url{https://commoncrawl.org/blog/news-dataset-available} \\
        RealNews \cite{RealNews} & CommonCrawl & \url{https://github.com/rowanz/grover/tree/master/realnews}\\
        RefinedWeb \cite{penedo2023refinedweb} & CommonCrawl & \url{https://huggingface.co/datasets/tiiuae/falcon-refinedweb}\\
        WebText & Reddit Link & Not open source yet \\
        OpenWebText \cite{Gokaslan2019OpenWeb} & Reddit Link & \url{https://skylion007.github.io/OpenWebTextCorpus/}\\
        PushShift.io \cite{baumgartner2020pushshift} & Reddit Link & \url{https://pushshift.io/}\\
        Wikipedia \cite{Wikipedia} & Wikipedia & \url{https://dumps.wikimedia.org/zhwiki/latest/}\\
        BigQuery \cite{bigquery-google} & Code & \url{https://cloud.google.com/bigquery}\\
        CodeParrot & Code & \url{https://huggingface.co/codeparrot} \\
        the Pile \cite{gao2020pile} & Other & \url{https://github.com/EleutherAI/the-pile}\\
        ROOTS \cite{roots} & Other & \url{https://huggingface.co/bigscience-data} \\
        \bottomrule
	\end{tabular}
\end{table}

\textbf{Books:} Two commonly utilized books datasets for LLMs training are BookCorpus \cite{BookCorpus} and Gutenberg \cite{Gutenberg}. These datasets include a wide range of literary genres, including novels, essays, poetry, history, science, philosophy, and more. Widely employed by numerous LLMs \cite{touvron2023llama,smith2022using}, these datasets contribute to the models' training by exposing them to a diverse array of textual genres and subject matter, fostering a more comprehensive understanding of language across various domains. 

\textbf{CommonCrawl:} CommonCrawl \cite{commoncrawl} manages an accessible repository of web crawl data, freely available for utilization by individuals and organizations. This repository encompasses a vast collection of data, comprising over 250 billion web pages accumulated over a span of 16 years. Established in 2007, Common Crawl has evolved into a widely recognized and referenced corpus in the academic and research communities, cited in more than 10,000 research papers. This continuously expanding corpus is a dynamic resource, with an addition of 3–5 billion new web pages each month. Its significance extends to the field of natural language processing, where it serves as a primary training corpus in numerous large language models. Notably, a substantial portion of the raw tokens employed in training GPT-3 \cite{GPT3}, amounting to 82\%, is sourced from the CommonCrawl. However, due to the presence of a substantial amount of low-quality data in web archives, preprocessing is essential when working with CommonCrawl data. Currently, four commonly used filtered datasets based on CommonCrawl are available: C4 \cite{T5}, CC-Stories \cite{CC-Stories}, CC-News \cite{liu2019roberta}, and RealNews \cite{RealNews}.

\textbf{Reddit Links:} Reddit is a social media platform where users can submit links and posts, and others can vote on them using the "upvote" or "downvote" system. This characteristic makes it a valuable resource for creating high-quality datasets.

\textbf{Wikipedia:} Wikipedia \cite{Wikipedia}, a free and open online encyclopedia project, hosts a vast repository of high-quality encyclopedic content spanning a wide array of topics. The English version of Wikipedia is extensively utilized in the training of many LLMs \cite{GPT3,touvron2023llama,thoppilan2022lamda}, serving as a valuable resource for language understanding and generation tasks. Additionally, Wikipedia is available in multiple languages, providing diverse language versions that can be leveraged for training in multilingual environments.

\textbf{Code:} There is a limited availability of publicly accessible code datasets at present. Existing efforts primarily involve web scraping of code with open-source licenses from the internet. The main sources include Github and Stack Overflow.

We have organized datasets utilized by distinct LLMs. During the training process, LLMs are typically trained on multiple datasets, as specified in Table~\ref{tab:llm_data} for reference.

\begin{table}[h]
	\caption{Datasets utilized by distinct LLMs}
    \label{tab:llm_data}
	\centering
	\begin{tabular}{cc}
	\toprule
	LLMs & Datasets \\
	\midrule	
	GPT-3 \cite{GPT3} & CommonCrawl \cite{commoncrawl}, WebText2 \cite{GPT3}, Books1 \cite{GPT3}, Books2 \cite{GPT3}, Wikipedia \cite{Wikipedia}\\
        LLaMA \cite{touvron2023llama} &  CommonCrawl \cite{commoncrawl}, C4 \cite{T5}, Wikipedia \cite{Wikipedia}, Github, Books, Arxiv, StackExchange\\
        PaLM \cite{chowdhery2022palm} &  Social Media, Webpages, Books, Github, Wikipedia, News (total 780B tokens)\\
        T5 \cite{T5} &  C4 \cite{T5}, WebText, Wikipedia, RealNews \\
        CodeGen \cite{Codegen}  & the Pile, BIGQUERY, BIGPYTHON \\
        CodeGeeX \cite{zheng2023codegeex} & CodeParrot, the Pile, Github\\
        GLM \cite{zeng2022glm}  &  BooksCorpus, Wikipedia\\
        BLOOM \cite{workshop2022bloom} & ROOTS\\
        OPT \cite{zhang2022opt} & BookCorpus, CCNews, CC-Stories, the Pile, Pushshift.io\\
        \bottomrule
	\end{tabular}
\end{table}

\subsubsection{Data preprocessing}
Once an adequate corpus of data is collected, the subsequent step is data preprocessing. The quality of data preprocessing directly impacts the model's performance and security. The specific preprocessing steps involve filtering low-quality text, including eliminating toxic and biased content to ensure the model aligns with human ethical standards. It also includes deduplication, removing duplicates in the training set, and excluding redundant content in the test set to maintain the sample distribution balance. Privacy scrubbing is applied to ensure the model's security, preventing information leakage or other privacy-related concerns. Additionally, if fine-tuning LLMs is considered, expanding the vocabulary should also be considered. 
On the other hand, LLaMA 2 models \cite{touvron2023llama2} represent a notable exception. These models forego filtering in their pretraining corpus, as aggressive filtration might accidentally filter out some demographic groups. This approach enhances the generalizability of the base LLaMA 2 models, making them more adept across a range of downstream tasks, such as hate speech detection and privacy de-identification. Observations indicate that abstaining from additional filtering in the pretraining data enables the base model to achieve reasonable safety alignment with fewer examples \cite{touvron2023llama2}. While this increases both generalizability and safety alignment efficiency, the implementation of additional safety mitigations is still imperative prior to public deployment, as further discussed in Section~\ref{sec::safety_ft}.

\textbf{Quality filtering:} Filtering low-quality data is typically done using heuristic-based methods or classifier-based methods. Heuristic methods involve employing manually defined rules to eliminate low-quality data \cite{Gopher,penedo2023refinedweb}. For instance, rules could be set to retain only text containing digits, discard sentences composed entirely of uppercase letters, and remove files with a symbol and word ratio exceeding 0.1, and so forth. Classifier-based methods involve training a classifier on a high-quality dataset such as WebText \cite{GPT2} to filter out low-quality datasets. 

\textbf{Deduplication:} Language models may sometimes repetitively generate the same content during text generation, potentially due to a high degree of repetition in the training data. Extensive repetition can lead to training instability, resulting in a decline in the performance of LLMs \cite{hernandez2022scaling}. Additionally, it is crucial to consider avoiding dataset contamination by removing duplicated data present in both the training and testing set \cite{lee2021deduplicating}.

\textbf{Privacy scrubbing:} LLMs, as text-generating models, are trained on diverse datasets, which may pose privacy concerns and the risk of inadvertent information disclosure \cite{carlini2021extracting}. In the preprocessing phase of language datasets, it is imperative to address privacy concerns by systematically removing any sensitive information. This involves employing techniques such as anonymization, redaction, or tokenization to eliminate personally identifiable details, geolocation, and other confidential data. By carefully scrubbing the dataset of such sensitive content, researchers and developers can ensure that the language models trained on these datasets uphold privacy standards and mitigate the risk of unintentional disclosure of private information. It is essential to strike a balance between data utility and privacy protection, fostering responsible and ethical use of language datasets in various applications.

\textbf{Filtering out toxic and biased text:} In the preprocessing steps of language datasets, a critical consideration is the removal of toxic and biased content to ensure the development of fair and unbiased language models. This involves implementing robust content moderation techniques, such as employing sentiment analysis, hate speech detection, and bias identification algorithms. By leveraging these tools \cite{gehman2020realtoxicityprompts}, researchers can systematically identify and filter out text that may perpetuate harmful stereotypes, offensive language, or biased viewpoints.

\subsection{Architecture}
Currently, all LLMs are built upon the Transformer architecture, allowing these models to scale to several 10 billion or even a trillion parameters. Typically, PLM architectures fall into three categories: Encoder-only \cite{devlin2018bert}, Encoder-decoder \cite{T5} and Decoder-only \cite{GPT}. The Encoder-only architecture is no longer employed in the latest LLMs and won't be further discussed here. Instead, this section will focus on introducing the Encoder-decoder and Decoder-only architectures.

\begin{figure}[h]
    \centerline{\includegraphics[width=\columnwidth]{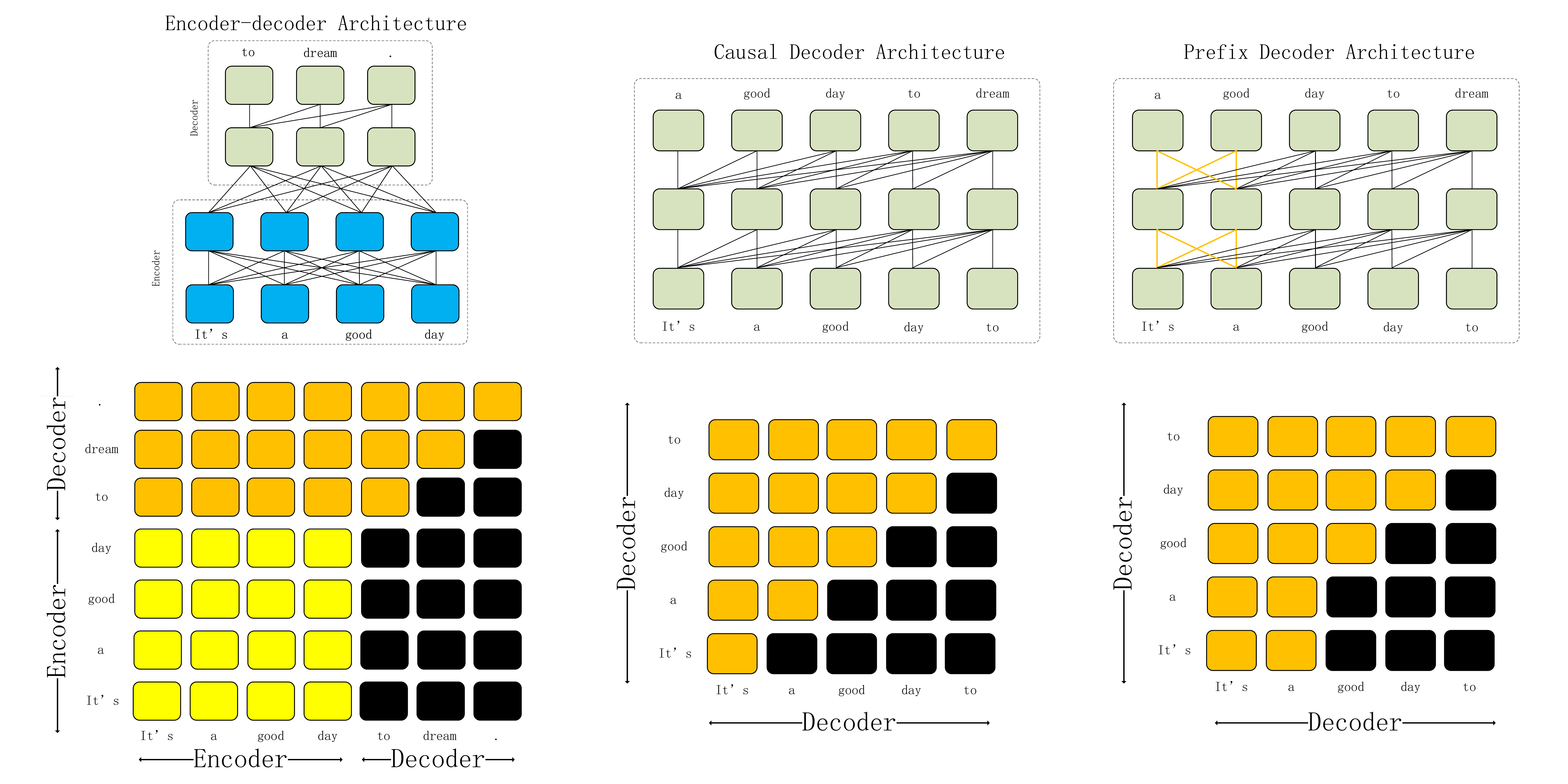}}
    \caption{The figures from left to right represent the Encoder-decoder architecture, Causal Decoder architecture, Prefix Decoder architecture, and their mask configurations, respectively. This diagram illustrates the range of tokens that each input token can attend to.} 
    \label{fig:architecture}
\end{figure}

\subsubsection{Encoder-decoder Architecture}

The Encoder-decoder architecture of LLMs is built upon the traditional Transformer Encoder-decoder architecture. The Encoder-decoder architecture consists of two main components: the Encoder and the Decoder. Each part of the Encoder is composed of multiple layers of Transformer's Multi-Head Self-Attention layers, which encode the input sequence. The Decoder, on the other hand, utilizes cross-attention over the output representation of the Encoder and generates the target sequence in an autoregressive manner. The encoder-decoder architecture serves as the foundation for prominent LLMs such as T5 \cite{T5}, flan-T5 \cite{flan-T5}, and BART \cite{lewis2019bart}.

\subsubsection{Decoder-only Architecture}
% GPT OPT BLOOM Gopher LLaMA
% PalM GLM prefix
LLMs with a Decoder-only architecture utilize the decoder component of the traditional Transformer architecture. Unlike the Encoder-Decoder architecture, which incorporates both an encoder and a decoder, the Decoder-only architecture is solely focused on the decoding process. In this configuration, the model sequentially generates tokens, attending to preceding tokens in the sequence. This architecture has been applied to various language generation tasks, showcasing its effectiveness in various tasks such as text generation without the need for an explicit encoding phase. The Decoder-only architecture can be further classified into two categories: the Causal Decoder architecture and the Prefix Decoder architecture. 

\textbf{The Causal Decoder Architecture: }In the Causal Decoder architecture, each token in the model input sequence can only attend to past input tokens and itself during the decoding process. It achieves unidirectional attention to the input sequence by using a specific mask as shown in Figure~\ref{fig:architecture}. In fact, different architectures are mainly implemented by configuring different mask matrices. The figure illustrates a comparison of mask configurations between the Encoder-decoder and Decoder-only architectures (including Casual Decoder and Prefix Decoder). The representative LLMs for the Causal Decoder architecture are the GPT series \cite{GPT,GPT-2,GPT3,instructGPT,openai2023gpt4}. The GPT series of LLMs are currently known for their superior performance, with their foundational Causal Decoder architecture widely applied in other LLMs such as BLOOM \cite{workshop2022bloom}, OPT \cite{zhang2022opt}, Gopher \cite{Gopher}, and LLaMA \cite{touvron2023llama}. 

\textbf{The Prefix Decoder Architecture: }The Prefix Decoder architecture combines the advantages of both the Encoder-decoder and Causal Decoder architectures. It leverages its unique mask configurations, as illustrated in Figure~\ref{fig:architecture}, enabling bidirectional attention for tokens in the prefix while maintaining unidirectional attention for generating subsequent tokens \cite{dong2019unified}. This design allows for the autoregressive generation of the output sequence with the flexibility to attend bi-directionally to the prefix tokens. Representative LLMs utilizing the Prefix Decoder architecture include PaLM \cite{chowdhery2022palm} and GLM \cite{zeng2022glm}.

\subsection{Pre-training Tasks}
Large Language Models (LLMs) typically learn rich language representations through a pre-training process. During pre-training, these models leverage extensive corpora, such as text data from the internet, and undergo training through self-supervised learning methods. Language modeling is one common form of self-supervised learning task in which the model is tasked with predicting the next word in a given context. Through this task, the model acquires the ability to capture information related to vocabulary, grammar, semantics, and text structure.

In language modeling \cite{GPT,GPT-2,GPT3,chowdhery2022palm}, the model is required to predict the next word in a given context. This task enables the model to develop a nuanced understanding of language. Specifically, the model observes large amounts of textual data and attempts to predict the next word at each position in the text. This gradual learning process allows the model to capture the patterns and information inherent in language, encoding a vast amount of linguistic knowledge into its parameters. Once pre-training is complete, these model parameters can be fine-tuned for various natural language processing tasks to adapt to specific task requirements. The objective of language modeling is to train a model to maximize the likelihood of textual data. For a given text sequence, denoted as $w_{1}, w_{2}, ..., w_{T}$, where $w_{t}$ represents the token at position $t$, $P(w_{t}|w_{1}, w_{2}, ..., w_{t-1})$  is the probability of predicting $w_{t}$ given the preceding context $w_{1}, w_{2}, ..., w_{t-1}$, the objective function for language modeling can be expressed using cross-entropy loss. Here, we define the objective as maximizing the conditional probability of the given text sequence:
\begin{equation}
    L_{LM} = \frac{1}{T} \sum_{t=1}^{T} -log P(w_{t}|w_{1}, w_{2}, ..., w_{t-1})
\end{equation}

Language modeling serves as a prevalent pretraining objective for most LLMs. In addition to language modeling, there are other pretraining tasks within the realm of language modeling. For instance, some models \cite{T5,zeng2022glm} use text with certain portions randomly replaced, and then employ autoregressive methods to recover the replaced tokens. The primary training approach involves the autoregressive recovery of the replaced intervals.

\subsection{Model Training}

\subsubsection{Parallel Training}
\begin{figure}[h]
    \centerline{\includegraphics[width=\columnwidth]{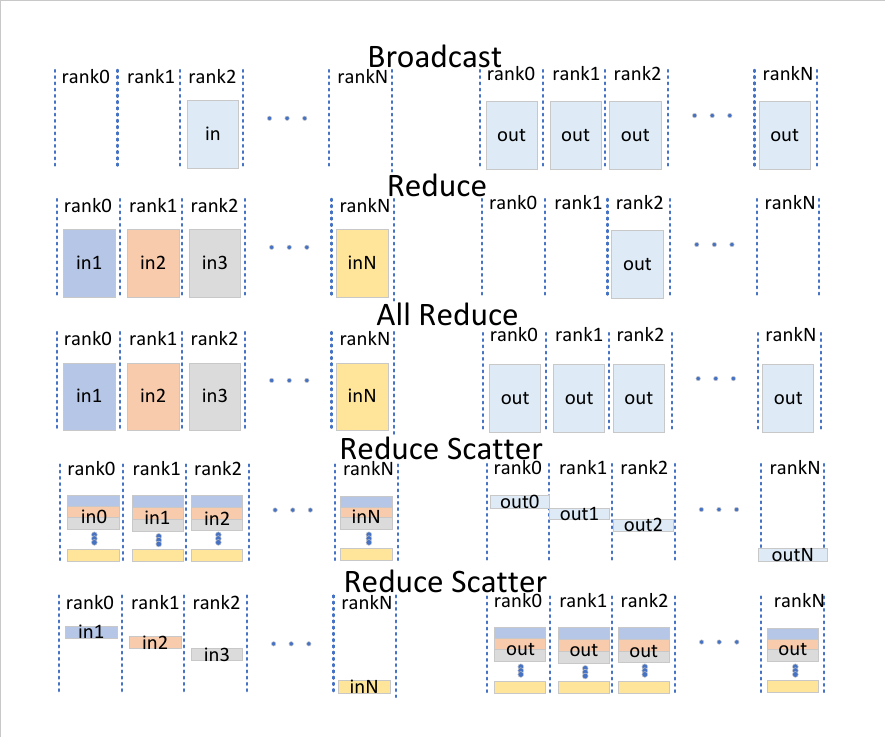}}
    \caption{Five collective communications that are used by parallel training methods.}
    \label{fig:collective-communication}
\end{figure}

In the parallel training mentioned below, there will be discussions about collective communication which helps us better understand the principles of parallel training. Figure~\ref{fig:collective-communication} has listed five reduction relationships. 1)Broadcast: Send data from one GPU to other GPUs.2)Reduce: Reduce(sum/average) data of all GPUs, send to one GPU.3)All  Reduce: Reduce all data of GPUs, send to all GPUs.4)Reduce Scatter: Reduce all data of GPUs, send portions to all GPUs.5)All Gather: Gather data of all GPUs, send all GPUs.

\begin{figure}[h]
    \centerline{\includegraphics[width=\columnwidth]{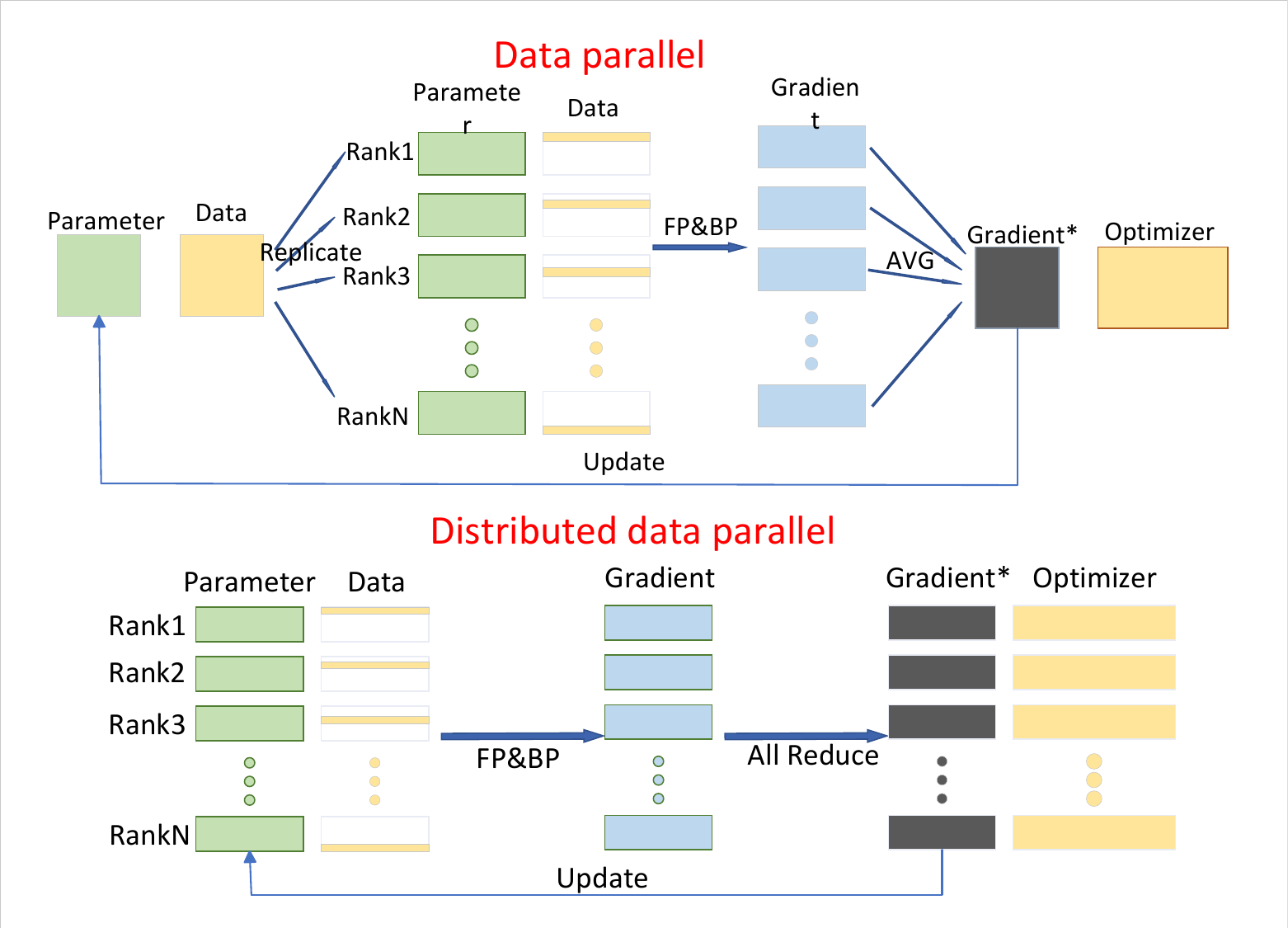}}
    \caption{The architecture of data parallelism and distributed data parallelism. The diagram illustrates the difference between data parallelism and distributed data parallelism and the advantages of distributed data parallelism.}
    \label{fig:dataparallelism}
\end{figure}

\textbf{Data Parallel:} The process of data parallelism \cite{li2020pytorch,} is shown in Figure~\ref{fig:dataparallelism}, there is a parameter server that stores the model's parameters and the entire batch of data. Each GPU uses broadcast to synchronize the model parameters and divides the data into one portion per GPU, with each GPU receiving a portion of the data. Each GPU uses the complete model parameters and a portion of the data to perform forward and backward propagation. This way, the gradients are obtained for each GPU. Finally, we aggregate the gradients and send the aggregated gradients back to the parameter server, where the original model parameters and the aggregated complete gradients are available. With this information, we can use an optimizer to update the model parameters. The updated parameters will then enter the next round of model training iterations.
Distributed data parallelism \cite{isard2007dryad} abandons the use of a parameter server and instead employs all-reduce on gradient information, ensuring that each GPU receives the same gradient information. The result of all-reduce is communicated to all GPUs, allowing them to independently update their respective model optimizers. After each round of updates, the model's parameters, gradients, and the historical information of the optimizer are consistent across all GPUs.

The occupation of GPU memory of intermediate results is related to the batch size, sentence length, and model dimensions. When using data parallelism, a batch of data is divided into many parts, allowing each GPU to process a portion of the data. In equivalent terms, the batch size processed on each GPU is reduced to one over the original number of GPUs. Data parallelism has reduced the input dimensions, resulting in an overall reduction in the intermediate results of the model. A drawback is that to support model training, each GPU needs to receive at least one piece of data. In the most extreme case, when each GPU receives only one piece of data, our parameters, gradients, and optimizer still need to be fully stored on the GPU. Even if we don't store any intermediate results on the GPU, our model may still be unable to perform computations on a single GPU.

\begin{figure}[h]
    \centerline{\includegraphics[width=\columnwidth]{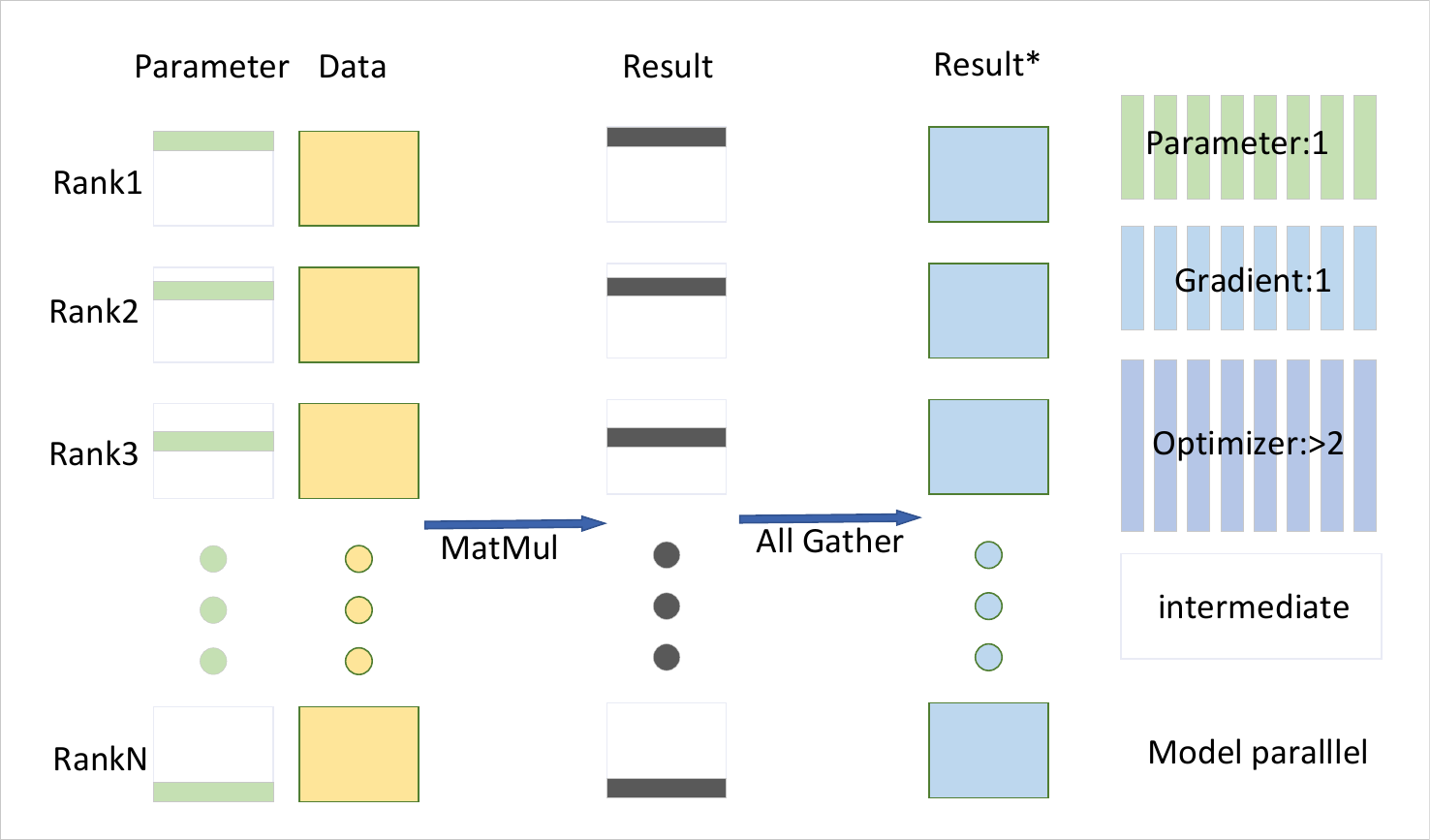}}
    \caption{The overall architecture of model parallelism. The left side of the diagram shows the process of model parallelism, and the right side shows the memory usage of parameters, gradients, and optimizers in the graphics card of the model parallelism method.}
    \label{fig:modelparallelism}
\end{figure}

\textbf{Model Parallel:}Model parallelism \cite{shoeybi2019megatron} was first introduced by Megatron-LM to alleviate memory pressure. From figure \ref{fig:modelparallelism}, we can clearly understand the overall architecture of model parallelism. Taking advantage of the most common linear layer in the Transformer as an example, the parameters of the linear layer form a matrix of size A*B, and the input to the linear layer is a vector of size B*1. Representing this as $y_{A*B}$ = $W_{A*B}$$x_B$, we can horizontally partition the model's parameters into many segments using the property of matrix multiplication. Each segment is of size a divided by n multiplied by B. Utilizing the properties of matrix multiplication, we can move $x_B$ into parentheses, and finally, the result of the linear layer is obtained by multiplying many small matrices with the parameters of the linear layer. Through this approach, the parameters of the linear layer can be distributed across multiple GPUs. However, it is crucial to ensure that the inputs to the model on multiple GPUs are identical. Instead of using a data parallel approach to partition the data, we need to ensure that the inputs obtained on each GPU are the same, meaning they belong to the same batch of data. We can then partition a parameter like the linear layer across GPUs, with each GPU receiving a small portion of the matrix. By performing model calculations with this small portion and the data, we obtain a sub-result, as shown in Formula 5. The results of these computations need to be concatenated using the all-gather operator and communicated to all GPUs.
\begin{equation}
  \begin{aligned}
    y_{A*B}&= W_{A*B}x_{B}\\       
           &=[W^{(1)}_{\frac{A}{n}*b};W^{(2)}_{\frac{A}{n}*b};...;W^{(n)}_{\frac{A}{n}*b}]x_{B}\\
           & = [W^{(1)}_{\frac{A}{n}*b}x_{B};W^{(2)}_{\frac{A}{n}*b}x_{B};...;W^{(n)}_{\frac{A}{n}*b}x_{B}]
  \end{aligned}    
\end{equation}

\begin{figure}[h]
    \centerline{\includegraphics[width=\columnwidth]{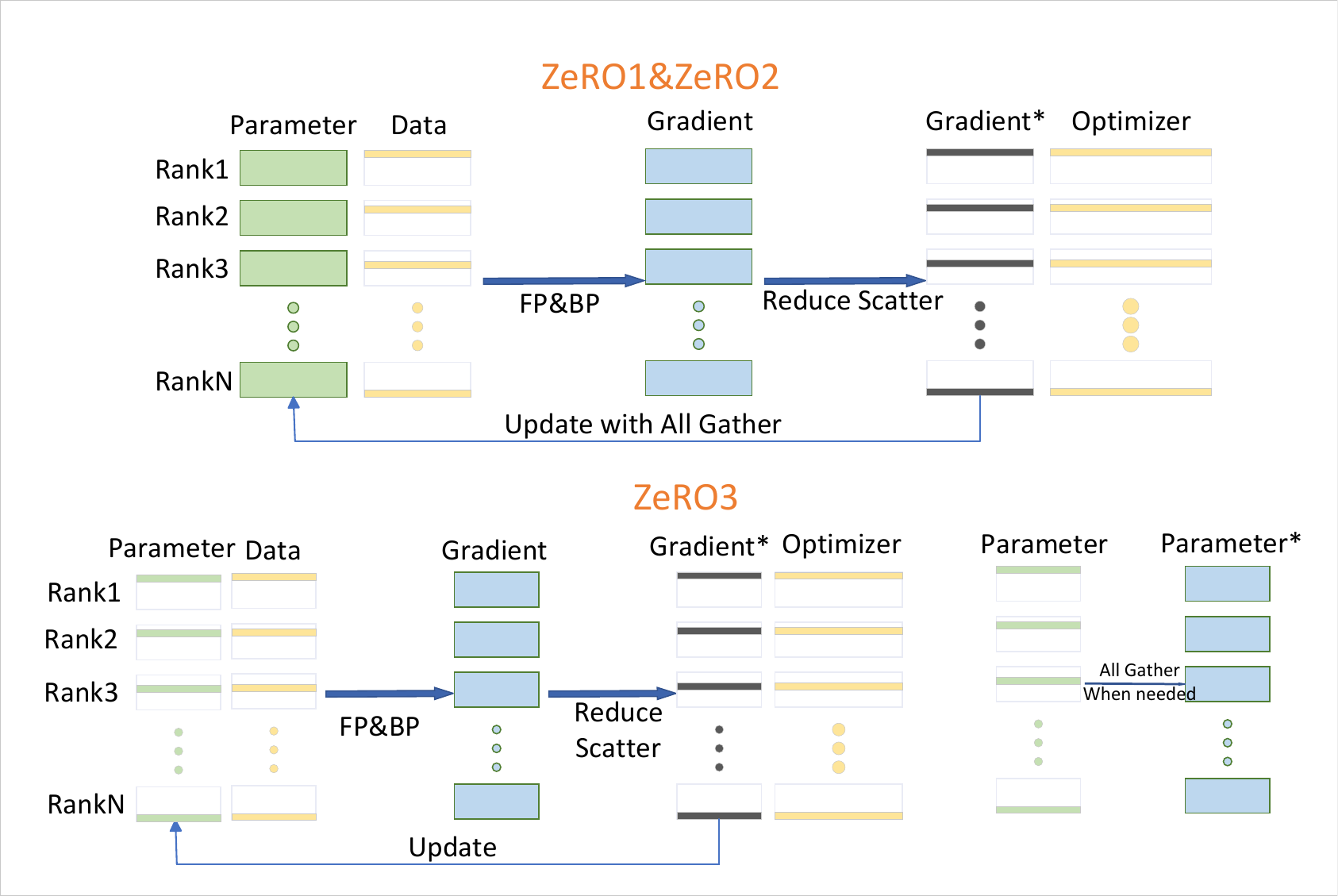}}
    \caption{The overall architecture of ZeRO. The upper demonstrates ZeRO stage1 and ZeRO stage2. The lower displays ZeRO stage3. The graph illustrates the optimization of memory usage of graphics card parameters in relation to ZeRO3 versus ZeRO1 and ZeRO2}
    \label{fig:ZeRO}
\end{figure}

\textbf{ZeRO:} ZeRO \cite{rajbhandari2020zero} is a framework built on data parallelism. During the parameter updating process on each GPU, the same set of parameters is used, leading to computational redundancy. Each GPU uses reduced scatter to eliminate this redundancy to obtain a portion of the gradient results. After updating a portion of the model parameters on each GPU, an all-gather operation is performed to synchronize the parameters across all GPUs. After the all-gather operation, the original gradient no longer needs to be saved on the graphics card and can be removed. Figure \ref{fig:ZeRO} shows the update of ZeRO. In ZeRO1, the original gradient is removed after backward propagation, while in ZeRO2, the product of the gradient* is calculated in advance during backward propagation, and only the gradient* is saved on the graphics card, removing the gradient. This way, the deletion of the gradient is advanced, leading to further savings in GPU memory space. ZeRO3 conducts a detailed division of the model parameters. Each graphics card retains only a portion of the gradients for updating, and parameter updates also only affect a portion of the model parameters. Therefore, each graphics card only needs to store the parameters, gradients, and optimizer related to the part of the parameters it is responsible for. During forward and backward propagation, an all-gather operation is required once, and after the operation is complete, the model parameters are released from the graphics card.Zero3 does not use all gather during parameter updates, but it requires an all-gather operation during both forward and backward propagation, adding one communication step. Compared to ZeRO2, ZeRO3 is an algorithm that trades time for space.

\textbf{Pipeline Parallel:} Pipeline parallelism \cite{huang2019gpipe} and model parallelism share similarities. In model parallelism, linear layers are divided into many small matrices, which are then distributed to different GPUs. For pipeline parallelism, different layers of the model are assigned to different GPUs. Specifically, if we have an n-layer transformer, we can assign the $layer_i$ of the transformer to the $GPU_i$, and so on. During the forward propagation of the model, we need to perform the computation of the $layer_i$ on the $GPU_i$, then pass the result to the $GPU_{i+1}$. The $GPU_{i+1}$ receives the output from the $GPU_i$, performs the computation for that layer and passes the result to the next GPU. This method partitions the parameters, gradients, optimizer, and intermediate results for each layer. 

\subsubsection{Mixed Precision Training}
In recent years, to pre-train extremely large language models, some research \cite{micikevicius2017mixed} has begun to utilize 16-bit floating-point numbers (FP16) to reduce memory usage and communication overhead. FP16 has a smaller numerical range and lower precision in effective digits \cite{rae2021scaling,workshop2022bloom}, but computations tend to be faster than FP32. In general model training, FP32 is often used as the default representation for training parameters. However, in actual model training, the number of parameters in a model typically does not exceed the order of thousands, well within the numerical range of FP16. To improve computational speed, we can convert from FP32 to FP16. During parameter updates, the amount of the parameter is roughly equal to the gradient multiplied by the learning rate. The minimum value of FP16 is on the order of 1e-5. As the product of the gradient and learning rate is already well below the representation range of FP16, the parameter update would result in loss, known as underflow. Therefore, we represent the parameter update obtained by multiplying the gradient by the learning rate as FP32. We cannot directly add this high-precision parameter update to a lower-precision model, as this would still result in floating-point underflow. Consequently, we need to save an additional single-precision parameter on the optimizer. To accelerate both forward and backward passes in the model, half-precision parameters and gradients are used and passed to the optimizer for updating. The optimizer's update quantity is saved as FP32, and we accumulate it effectively through a temporarily created FP32 parameter in the optimizer. After effective accumulation, it is then converted back to FP16 parameters.

\subsubsection{Offloading}
The parameters in the optimizer are at least twice as many as the model parameters, and a study \cite{ren2021zero}proposes the idea of moving the optimizer's parameters from the GPU to the CPU. Although GPU computation is much faster than CPU, the question arises whether offloading this operation could become a bottleneck for the overall training speed of the model optimizer. In reality, we utilize ZeRO3. After the optimization with ZeRO3, the size of the parameters, gradients, and optimizer is reduced to 1/n of the number of GPUs. By binding one GPU to multiple CPUs, we effectively lower the computational load on each CPU.

\subsubsection{Overlapping} 
Memory operations are typically asynchronous. Thus, We can send a request to the memory in advance and then proceed with other computations. After completing other computations, we come back to handle the memory request. This two-step operation is used in the forward propagation process of model training. We need to obtain the parameters of $layer_i$ through a gather operation. After obtaining the parameters of $layer_i$, in the forward propagation process of $layer_i$, we proactively retrieve the parameters of $layer_{i+1}$ through an asynchronous fetch. Once the forward propagation calculation for $layer_i$ is completed, the parameters for $layer_{i+1}$ have been obtained and are stored in the GPU. We can then immediately proceed with the forward propagation calculation, and so on.

\subsubsection{Checkpoint}
In order to support the backward propagation of the model, All intermediate results in the GPU memory need to be saved during the forward propagation of the model. To optimize this process, a checkpoint mechanism, which does not save all intermediate results in the GPU memory but only retains certain checkpoint points is utilized.

The diagram below illustrates a simplified structure of a transformer. Each transformer block takes a model input, undergoes complex computations through attention and feed-forward processes, and produces the overall output of that layer. We keep only the input of each major layer in the transformer as our checkpoint.

During the backward propagation process, how do we compute the gradients of the linear layers within each major layer? We can perform a technique called recomputation, which involves re-executing the forward pass of each major layer during the backward propagation process. We temporarily obtain the inputs of the linear layers within each major layer, and the intermediate results obtained can be used for backward propagation. Once the backward propagation for that layer is complete, we can discard the checkpoint and the temporarily recomputed intermediate results of the linear layers within the model from the GPU memory.

Assuming we have a transformer with 24 layers, each layer containing four to five linear layers, using the checkpoint mechanism reduces the originally required storage of 120 intermediate results to only 24 intermediate results.

\subsection{Fine-Tuning}
\label{sec::fine}
The training of LLMs in this paper is divided into three stages: data collection and processing, pre-training, and fine-tuning. This section will provide a review of the fine-tuning methods for LLMs. Specifically, we categorize fine-tuning techniques into three types: supervised fine-tuning (SFT) \cite{instructGPT}, alignment tuning, and parameter-efficient tuning.

\subsubsection{Supervised Fine-Tuning}
The core concept of supervised fine-tuning involves adjusting the model in a supervised manner on the basis of large-scale pre-training, enhancing its capability to better adapt to the specific requirements of the target task. In the process of SFT, it is necessary to prepare a labeled dataset for the target task, which includes input text along with corresponding labels. Instruction tuning is a commonly used technique in the fine-tuning process of LLMs and can be considered as a specific form of SFT. It involves further training LLMs on a dataset composed of (instruction, output) pairs, focusing on enhancing the capabilities and controllability of large language models by understanding and following human instructions. We compiled commonly used instruction tuning datasets, as illustrated in Table~\ref{tab:instruction}.

\begin{table}[h]
	\caption{Commonly used instruction tuning datasets.}
    \label{tab:instruction}
	\centering
	\begin{tabular}{cc}
	\toprule
	Datasets & Links \\
	\midrule	
	static-hh & \url{https://huggingface.co/datasets/Dahoas/static-hh} \\
        OIG & \url{https://huggingface.co/datasets/laion/OIG} \\
        Self-Instruct \cite{wang2022self} & \url{https://github.com/yizhongw/self-instruct} \\
        Natural instructions \cite{wang2022super}& \url{https://github.com/allenai/natural-instructions}\\
        P3 \cite{bach2022promptsource} & \url{https://huggingface.co/datasets/bigscience/P3}\\
        Promptsource \cite{victor2022multitask} & \url{https://github.com/bigscience-workshop/promptsource}\\
        WebGPT \cite{nakano2021webgpt}& \url{https://huggingface.co/datasets/openai/webgpt_comparisons}\\
        Flan \cite{weifinetuned} & \url{https://github.com/google-research/flan}\\
        MVPCorpus \cite{tang2022mvp} & \url{https://github.com/RUCAIBox/MVP}\\
        \bottomrule
	\end{tabular}
\end{table}

\subsubsection{Alignment Tuning}
Due to LLMs being pre-trained on massive and diverse internet data, even though the training data undergoes some preprocessing, it is still challenging to guarantee the absence of biased or harmful content in terabyte-scale training datasets. Despite LLMs demonstrating impressive performance across various natural language processing tasks, they frequently exhibit behaviors diverging from human intent. This includes generating false information, producing expressions with bias or misleading content, and so on \cite{instructGPT,kenton2021alignment}. To address these issues of LLMs displaying behaviors beyond human intent, alignment tuning becomes crucial \cite{instructGPT,glaese2022improving}.

In general, alignment tuning aims to meet the following three criteria: being helpful, honest, and harmless.

\textbf{Helpful:} The concept of helpfulness revolves around whether the model-generated output proves genuinely beneficial for a specific task or inquiry. In the realm of natural language processing, the model's generated text or responses should furnish valuable information, positively impacting the user's requirements or task objectives.

\textbf{Honest:} Honesty entails whether the model-generated output is authentic and reliable. The model should produce information consistent with facts, steering clear of fabrication or distortion. This contributes to maintaining user trust in the authenticity of the model's outputs.

\textbf{Harmless:} Harmlessness is concerned with whether the model-generated output poses no harm to users or society. The model should refrain from generating content that is harmful, offensive, or perilous, ensuring its utilization remains safe for all relevant stakeholders.

In training LLMs, a noteworthy approach to alignment tuning is based on Reinforcement Learning with Human Feedback (RLHF) \cite{instructGPT}. This method involves collecting human feedback data to train a reward model (RM) for reinforcement learning. The RM serves as the reward function during reinforcement learning training, and algorithms such as Proximal Policy Optimization (PPO) \cite{schulman2017proximal} are employed to fine-tune the LLM. In this context, LLM is considered as the policy, and the action space is considered as the vocabulary of the LLM.

\subsubsection{Parameter-efficient Tuning}
Currently, large-scale PLMs such as ChatGPT \cite{instructGPT,openai2023gpt4} continue to grow in scale. However, for the majority of researchers, conducting full fine-tuning on consumer-grade hardware has become cost-prohibitive and impractical. Unlike SFT and alignment tuning, the objective of parameter-efficient tuning is to reduce computational and memory overhead. This method involves fine-tuning only a small or additional subset of model parameters while keeping the majority of pre-trained parameters fixed, thereby significantly lowering computational and storage costs. It is noteworthy that state-of-the-art parameter-efficient tuning techniques have achieved performance levels comparable to full fine-tuning. Some common parameter-efficient tuning methods include Low-Rank Adaptation (LoRA) \cite{hu2021lora}, Prefix Tuning \cite{li2021prefix} and P-Tuning \cite{liu2021p,liu2023gpt}. The adoption of these methods enables efficient model tuning even in resource-constrained environments, offering feasibility and efficiency for practical applications. 

With the rise of LLMs, parameter-efficient tuning has garnered increasing attention, with LoRA being widely employed in the latest releases of LLMs. LoRA \cite{hu2021lora} and its related advancements \cite{zhang2023adaptive,dettmers2023qlora} are noteworthy and deserve attention.

\subsubsection{Safety Fine-Tuning}
\label{sec::safety_ft}
To enhance the safety and responsibility of LLMs, the integration of additional safety techniques during fine-tuning is essential. This encompasses three primary techniques, applicable to both SFT and RLHF phases.

\textbf{Supervised Safety Fine-Tuning}: In this technique, labelers are tasked with generating demonstration data that incorporates high safety risk adversarial prompts. This handcraft safety demonstration data is then incorporated into the SFT phase, thereby augmenting the model's capacity to manage safety risks.

\textbf{Safety RLHF}: This technique employs the same or even more aggressive adversarial prompts to query the models. The safest response exhibiting refusal behavior is then used to train a safety reward model within the RLHF framework.

\textbf{Safety Context Distillation}: This technique employs context distillation \cite{askell2021general} by initially prefixing safety preprompts, like “You are a safe and responsible assistant,” to adversarial prompts. This process yields safer generated responses. The model is then fine-tuned on these safer demonstration data but without the inclusion of the safety pre-prompts. This safety distillation further enhances the model's safety capabilities.

\subsection{Evaluation}
Unlike in the past, large-scale deep learning models have a wider range of applications and stronger performance compared to ordinary models. However, with great power comes great responsibility, and evaluating these models has become more complex, requiring consideration of potential problems and risks from all aspects. Since the popularity of ChatGPT, many related studies have been published, including the survey and summary of LLMs evaluation in reference \cite{Chang2023ASO,liu2023holistic}, which is helpful for developing large-scale deep learning models. This section will introduce some testing datasets, evaluation directions and methods, and potential threats that need to be considered based on previous evaluation work on large models.

\subsubsection{Static testing dataset}
The evaluation of large models' capabilities requires appropriate datasets for validation. Here, we introduce several commonly used datasets for testing purposes. Considering multimodal large models, typical datasets for computer vision include ImageNet \cite{5206848} and Open Images \cite{Kuznetsova2020The}. In addition to the commonly used GLUE \cite{Wang2018GLUEAM} and SuperGLUE \cite{Wang2019SuperGLUEAS} for LLMs, MMLU \cite{Hendrycks2020MeasuringMM} is highly competitive in testing comprehensive capability. If your model primarily uses Chinese language, then CMMLU \cite{Li2023CMMLUMM}, as a benchmark for Chinese large models, should also be considered, and XTREME \cite{Hu2020XTREMEAM} and XTREME-R \cite{Ruder2021XTREMERTM} are suitable choices for multilingual large models. For assessing mathematical knowledge capabilities, there are datasets such as MATH \cite{Hendrycks2021MeasuringMP} and GSM8K \cite{Cobbe2021TrainingVT}, while HumanEval \cite{Chen2021EvaluatingLL} and MBPP \cite{Austin2021ProgramSW} can serve as benchmarks for code generation. For common sense reasoning tests in daily human life and work, the following datasets are available: HelloSwag \cite{Zellers2019HellaSwagCA}, PIQA \cite{Bisk2019PIQARA}, BoolQ \cite{Clark2019BoolQET}, SIQA \cite{2019SocialIQA}, WinoGrande \cite{Sakaguchi2019WinoGrande}, ARC \cite{Clark2018ThinkYH}, and OpenBookQA \cite{Mihaylov2018CanAS}. For medical knowledge, there are datasets such as MedQA-USMLE \cite{Jin2020WhatDD} and MedMCQA \cite{Pal2022MedMCQAA}.

\subsubsection{Open domain Q\&A evaluation}
Currently, LLMs interact with humans in the form of questions and answers. Compared to the fragmented and ambiguous information returned by traditional searches, LLMs provide more realistic and efficient question-and-answer results that align with human habits. Therefore, the evaluation of ODQA (Open Domain Question Answering) \cite{Voorhees1999TheTQ} capability is essential. The performance of open-domain question answering greatly affects user experience. Commonly used datasets for testing include SquAD \cite{Rajpurkar2016SQuAD1Q} and Natural Questions \cite{Kwiatkowski2019NaturalQA}, with F1 score and Exact-Match accuracy (EM) as evaluation metrics. However, note that the method of word matching may have certain issues, such as when a factually correct answer is not in the golden answer list. Therefore, human evaluation seems to be necessary, and literature \cite{Kamalloo2023EvaluatingOQ} has conducted detailed research on this matter.

\subsubsection{Security evaluation}
As an emerging and hot research field, LLMs must pay attention to their potential security threats, prevent malicious use or vulnerabilities to malicious attacks, and address any potential long-term issues that may pose a threat to human development. Additionally, red teaming in various domains is necessary to critically assess and test the model, identifying vulnerabilities, biases, inaccuracies, and areas for safety improvement.

\textbf{Potential bias:}The training data for LLMs may contain potential biases, such as gender or race. Security assessments need to address whether the model generates or amplifies these biases and how to reduce or correct them. Reference \cite{Ferrara2023ShouldCB} discusses in detail the causes of bias in LLMs and the serious consequences that may arise. Reference \cite{Gehman2020RealToxicityPromptsEN} extensively studies how pre-trained language models generate harmful content to what extent, and how to use controlled text generation algorithms to prevent the generation of toxic content. CHBias \cite{Zhao2023CHBiasBE} is a Chinese dataset that can be used to evaluate and mitigate the bias problem of LLMs.

\textbf{Privacy protection:} LLMs may come into contact with a large amount of user data, such as text and images, during the training process. Security assessments need to ensure the effective protection of user data privacy to prevent leaks and misuse. Reference \cite{Nasr_Carlini_Hayase_Jagielski_Cooper} conducted research on models like ChatGPT and found that it is possible to extract training data effectively from these models. Reference \cite{Wu2023DEPNDA} provides a solution by proposing a framework called DEPN (Detect and Editing Privacy Neurons) to detect and edit privacy neurons in pre-trained language models. It also introduces a privacy neuron aggregator to eliminate privacy information in a batch-processing manner, effectively reducing the leakage of privacy data while maintaining model performance.

\textbf{Adversarial attacks:} LLMs may be susceptible to adversarial attacks, such as input tampering, intentional misinformation, or generating false information. Security assessments need to consider the robustness of the model, i.e., its ability to withstand such attacks. As mentioned in reference \cite{Zou2023UniversalAT}, LLMs still have "jailbreak" risks, where users can manipulate the model to generate toxic content using specific input methods like role-playing or adding special suffixes as studied in the referenced paper. Especially when using open-source pre-trained models, any vulnerabilities in the pre-training models regarding adversarial attacks are inherited as well. Reference \cite{Zhang2022ReMoSRD} provides a solution to mitigate the harm caused by these vulnerabilities.

\subsubsection{Evaluation method}
Automated evaluation and manual evaluation play crucial roles in Language Model (LLM) research. Automated evaluation typically involves using various metrics and indicators to quantify the performance of models, such as BIEU \cite{Papineni2002BleuAM}, ROUGE \cite{Lin2004ROUGEAP}, and BERTSScore \cite{Zhang2019BERTScoreET}, which can measure the accuracy of LLM-generated content. These metrics can help researchers quickly assess model performance on large-scale data and compare different models. However, automated evaluation also has limitations as it cannot fully capture the complexity of language understanding and generation. Research in reference \cite{Novikova2017WhyWN} has shown that manual evaluation is more reliable for some open-ended generation tasks. Manual evaluation typically involves human annotators subjectively judging and assessing the quality of model-generated outputs. This evaluation method can help reveal how models perform in specific tasks or scenarios and identify subtle issues and errors that automated evaluation may overlook. However, manual evaluation also faces challenges such as high time costs and subjectivity. Therefore, it is often necessary to combine the strengths of automated and manual evaluation to comprehensively assess the performance of language models.

\subsection{LLM Framework}
Large deep learning models offer significant accuracy gains, but training billions to trillions of parameters is challenging. Existing solutions such as distributed training have solved fundamental limitations to fit these models into limited device memory while obtaining computation, communication, and development efficiency. Next, this section will introduce several large language model frameworks that utilize distributed training technology leveraging GPU, CPU, and NVMe memory to allow for unprecedented model scale on limited resources without requiring model code refactoring.

\textbf{Transformers}
Transformers\cite{wolf2020transformers}, an open-source Python library by Hugging Face, is dedicated to building models using the Transformer architecture. Featuring a simple and user-friendly API, it facilitates easy customization of various pre-trained models. With a robust community of users and developers, transformers continuously update and improve models and algorithms. 

\textbf{DeepSpeed: }Deepspeed \cite{rasley2020deepspeed}, an open-source optimization library compatible with PyTorch, is developed by Microsoft and utilized in training LLMs like MTNLG \cite{smith2022using} and BLOOM \cite{workshop2022bloom}. Currently, It provides full support for ZeRO technology which includes Optimizer state partitioning, Gradient partitioning and parameter partitioning, Custom mixed precision training, A range of fast CUDA-extension-based optimizers \cite{rajbhandari2021zero} and ZeRO-offload to CPU and Disk/NVMe. Through the above technologies. Additionally, Deepspeed has achieved excellent scalability and efficiency with small memory requirements.

\textbf{BMTrain: }BMTrain \cite{zeng2023openbmb} is an efficient large model training toolkit developed by Tsinghua University that can be used to train large models with tens of billions of parameters. It can train models in a distributed manner while keeping the code as simple as stand-alone training. BMTrain does not require model refactoring to work. In fact, PyTorch users can enable BMTrain with a few lines of code change to their existing training pipeline. It provides the support of various optimization techniques such as ZeRO optimization and communication optimization.

\textbf{Megatron-LM:} Megatron-LM \cite{shoeybi2019megatron,narayanan2021efficient,korthikanti2023reducing} is a deep learning library developed by NVIDIA for training large-scale language models.Megatron-LM presents their techniques including model and data parallelism, mixed-precision training, and FlashAttention for training very large transformer models. Specifically, it takes advantage of the structure of transformer networks to create a simple model parallel implementation by adding a few synchronization primitives and it enables training transformer models with billions of parameters and trains efficiently in PyTorch.It also performs an in-depth empirical analysis of their model and data parallel technique and demonstrates up to 76\% scaling efficiency using 512 GPUs which can largely improve the training efficiency and speed, enabling efficient distributed training across GPUs.

In addition to the aforementioned frameworks, Colossal-AI \cite{Colossal-ai} and FastMoE \cite{he2021fastmoe,he2022fastermoe} are also two popular frameworks for training LLMs. In principle, any deep learning framework that supports parallel computing can be used to train LLMs. Examples include PyTorch \cite{paszke2019pytorch}, TensorFlow \cite{abadi2016tensorflow,abadi2016tensorflow2}, PaddlePaddle \cite{ma2019paddlepaddle}, MXNet \cite{chen2015mxnet}, OneFlow \cite{yuan2021oneflow}, MindSpore \cite{huawei2022huawei} and JAX \cite{bradbury2018jax}. 

\section{Inference with Large Language Models}
\label{sec::inference}
The scale of large models is growing at a rate of nearly 10 times per year, which brings about huge computational consumption and carbon emissions \cite{strubell2019energy}. Therefore, reducing the computational burden of training large models while retaining their reasoning ability has become a common concern for everyone. In this chapter, we mainly introduce how to reduce costs from both computational and storage aspects, that is, how to efficiently perform large-scale model inference from four aspects: model compression, memory scheduling, parallelism, and structural optimization.
\subsection{Model Compression}

\subsubsection{Knowledge Distillation}
Knowledge Distillation \cite{hinton2015distilling} refers to transferring knowledge from a cumbersome (teacher) model to a smaller (student) model that is more suitable for deployment. This is achieved by fitting the soft targets of the two models, as soft targets provide more information than gold labels. Initially, the calculation for model distillation involved only fitting the outputs from the last layer of both the teacher and student models \cite{gou2021knowledge}. PKD \cite{sun2019patient} improves this process by computing the mean-square loss between normalized hidden states, allowing the student model to learn from multiple intermediate layers of the teacher model. In order to discover more intermediate representations suitable for knowledge distillation, Jiao et al. \cite{jiao2019tinybert} proposed Tiny BERT. This enables the student model to learn from the embedding layer and attention matrices of the teacher model.

\subsubsection{Model Pruning}
Model pruning involves removing redundant portions from the parameter matrices of large models. It is divided into unstructured pruning and structured pruning. Unstructured pruning involves removing individual connections or weights in a neural network without adhering to any specific structural pattern. In structured pruning, specific structural patterns or units within a neural network are pruned or removed. Gordon et al. \cite{gordon2020compressing} compared the effects of unstructured and structured pruning on the BERT model. They found that the effectiveness of unstructured pruning significantly decreases as the pruning ratio increases, while in structured pruning, 30-40\% of the weights can be discarded without affecting BERT's universality. 
Different structures in the model can be structurally pruned. Michel et al. \cite{michel2019sixteen} pruned attention heads and found that ablating one head often positively impacts the performance of WMT and BERT. They proposed a gradient-based metric for evaluating the importance of attention heads to enhance pruning effectiveness. Fan et al. \cite{gordon2020compressing} performed layer pruning by extending dropout from weights to layers. During training, they randomly dropped layers and achieved good inference results by selecting sub-networks with any desired depth during testing.

\subsubsection{Model Quantization}
The fundamental idea behind model quantization is to reduce the number of floating-point bits used in numerical calculations within a large model network, thereby decreasing storage and computation costs. This involves converting floating-point operations into fixed-precision operations. However, as precision decreases, the model's loss gradually increases, and when precision drops to 1 bit, the model's performance experiences a sudden decline. To address the optimization challenges introduced by low-precision quantization, Bai et al. \cite{bai2020binarybert} proposed BinaryBERT. They initially trained a half-sized ternary model and then initialized a binary model with the ternary model through weight splitting. Finally, they fine-tuned the binary model. This approach yielded better results for the binary model compared to training a binary model from scratch.

\subsubsection{Weight Sharing}
The basic idea of weight sharing is to use the same set of parameters for multiple parts of a LLM. Instead of learning different parameters for each instance or component, the model shares a common set of parameters across various parts. Weight sharing helps reduce the number of parameters that need to be learned, making the model more computationally efficient and reducing the risk of overfitting, especially in situations where there is limited data. ALBERT \cite{lan2019albert} uses the Cross-layer parameter-sharing strategy to effectively reduce the number of parameters of the model, and can achieve better training results than the baseline with the same parameter number.

\subsubsection{Low-rank Approximation}
Low-rank decomposition methods are crucial in the field of model compression, as they allow for the creation of more compact models with fewer parameters. This reduction in model size is particularly beneficial for deploying neural networks on resource-constrained devices, improving efficiency during inference. Chen et al. \cite{chen2021drone} performed a low-rank decomposition on the input matrix, enabling matrix operations within the large model to occur at a lower-rank level, effectively reducing the computational workload. From the results, their proposed method, DRONE, not only ensures the inference performance of the large model but also achieves an acceleration ratio of more than 1.3 times compared to the baseline method. The specific choice of low-rank decomposition method depends on the architecture of the neural network and the requirements of the target application.

\subsection{Memory Scheduling}
Deploying LLMs on a single consumer-grade GPU is constrained by the limitations of the available video memory, given the substantial parameters of LLMs. Therefore, appropriate Memory Scheduling strategies can be used to solve the hardware limitations of large model inference. Memory scheduling in large model inference involves the efficient organization and management of memory access patterns during the reasoning or inference phase of complex neural network models. In the context of sophisticated reasoning tasks, such as natural language understanding or complex decision-making, large models often have intricate architectures and considerable memory requirements. Memory scheduling optimizes the retrieval and storage of intermediate representations, model parameters, and activation values, ensuring that the inference process is both accurate and performed with minimal latency. For example, BMInf \cite{han2022bminf} utilizes the principle of virtual memory, achieving efficient inference for large models by intelligently scheduling the parameters of each layer between the GPU and CPU.

\subsection{Parallelism}
Both inference and training can leverage parallelization techniques. Presently, parallelization techniques for inference primarily manifest across three dimensions: Data Parallelism, Tensor Parallelism, and Pipeline Parallelism. Data Parallelism primarily involves increasing the overall throughput of the inference system by adding more GPU devices \cite{ren2021zero,rajbhandari2020zero,rajbhandari2021zero,zhao2023pytorch}. Tensor parallelism is a form of model parallelism where the model's parameters are partitioned into multiple tensors, each computed on different processing units. This approach proves beneficial when dealing with models that are too large to fit into the memory of a single GPU. Tensor parallelism primarily involves increasing the number of devices horizontally through parallel computation to reduce latency \cite{shoeybi2019megatron}. Pipeline parallelism primarily involves vertically increasing the number of GPU devices through parallel computation to support larger models and enhance device utilization. Typically, it is combined with tensor parallelism to achieve optimal performance \cite{huang2019gpipe}.

\subsection{Structural Optimization}
In the forward propagation computation of LLMs, the calculation speed is significantly faster than the speed of memory access. Inference speed can be impacted by numerous memory access operations. One goal in LLM inference is to minimize the number of memory accesses during forward propagation. FlashAttention \cite{dao2022flashattention} and PagedAttention \cite{kwon2023efficient} enhance computational speed by employing a chunked computation approach, mitigating the storage overhead associated with matrices. The entire operation takes place within SRAM, reducing the number of accesses to High Bandwidth Memory (HBM) and significantly boosting computational speed. Both FlashAttention and PagedAttention have been adopted by mainstream inference frameworks, and seamlessly integrated into these frameworks for straightforward utilization.

\subsection{Inference Framework}
Parallel computing, model compression, memory scheduling, and specific optimizations for transformer structures, all integral to LLM inference, have been effectively implemented in mainstream inference frameworks. These frameworks furnish the foundational infrastructure and tools required for deploying and running LLM models. They offer a spectrum of tools and interfaces, streamlining the deployment and inference processes for researchers and engineers across diverse application scenarios. The choice of a framework typically hinges on project requirements, hardware support, and user preferences. In Table~\ref{tab:Inference_framework}, we compile some of these frameworks for reference.

\begin{table}[h]
	\caption{List of LLM inference framework.}
    \label{tab:Inference_framework}
	\centering
	\begin{tabular}{cc}
	\toprule
	Framework & Links \\
	\midrule	
	TensorRT & \url{https://github.com/NVIDIA/TensorRT-LLM} \\
        FasterTransformer & \url{https://github.com/NVIDIA/FasterTransformer}\\
        Megatron-LM \cite{shoeybi2019megatron} & \url{https://github.com/NVIDIA/Megatron-LM}\\
        FlexGen \cite{sheng2023flexgen} & \url{https://github.com/FMInference/FlexGen} \\
        DeepSpeed \cite{rasley2020deepspeed} & \url{https://github.com/microsoft/DeepSpeed}\\
        vLLM \cite{kwon2023efficient} & \url{https://github.com/vllm-project/vllm}\\
        FlexFlow \cite{miao2023specinfer} & \url{https://github.com/flexflow/FlexFlow}\\
        StreamingLLM \cite{xiao2023efficient} & \url{https://github.com/mit-han-lab/streaming-llm}\\
        ColossalAI \cite{Colossal-ai}& \url{https://github.com/hpcaitech/ColossalAI} \\
        BMCook \cite{zhang2022bmcook} & \url{https://github.com/OpenBMB/BMCook}\\
        BMInf \cite{han2022bminf} & \url{https://github.com/OpenBMB/BMInf}\\
        Petals \cite{borzunov2022petals} & \url{https://github.com/bigscience-workshop/petals} \\
        \bottomrule
	\end{tabular}
\end{table}
\section{Utilization of LLMs}
\label{sec::app}

\begin{figure}[h]
    \centerline{\includegraphics[width=\columnwidth]{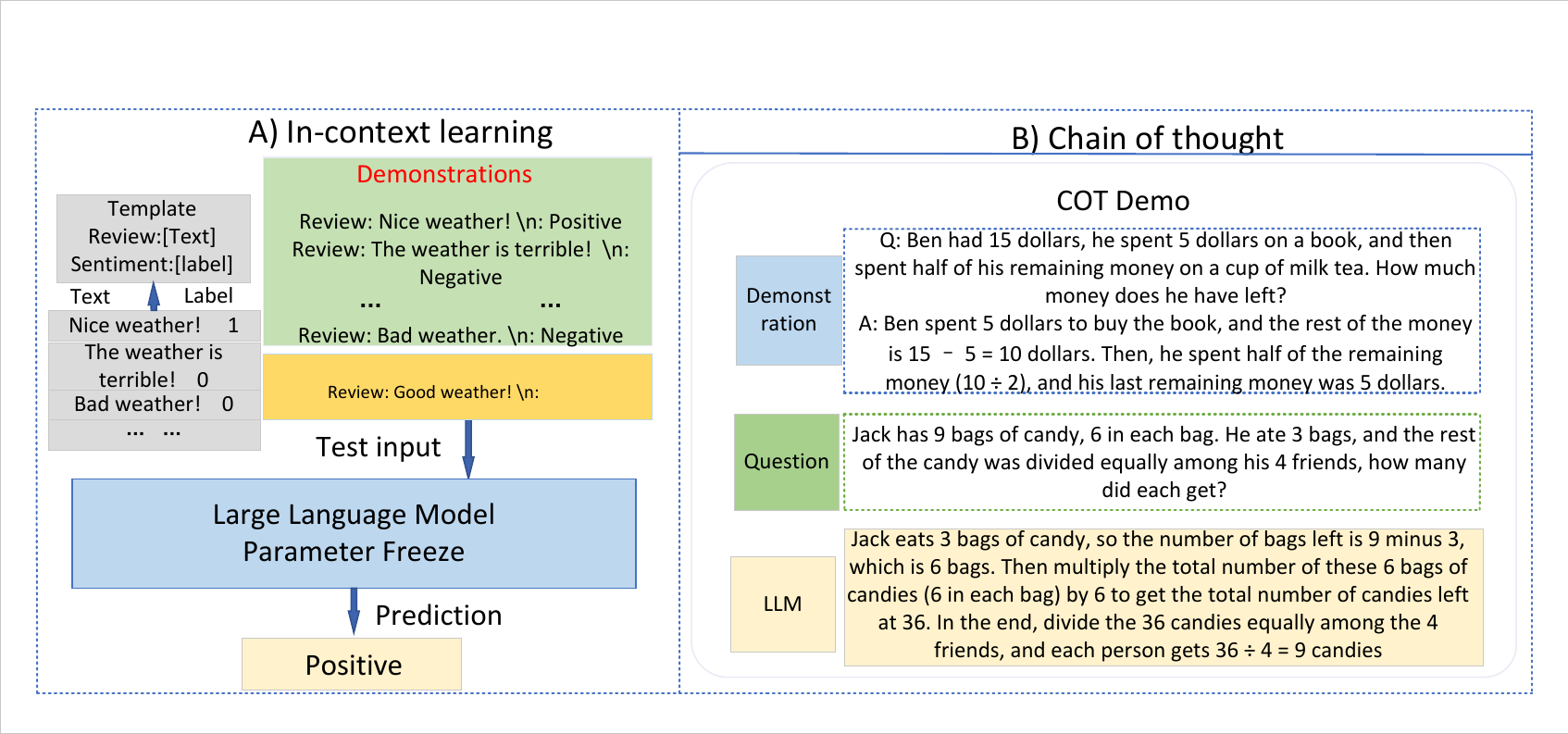}}
    \caption{A) in-context learning, B) Chain of thought.}
    \label{fig:TheUltilizationofLLM}
\end{figure}

The application scope of LLMs is extensive and can be practically employed in almost any specialized domain \cite{liu2023summary,dou2023artificial,li2023artificial,LIU2023100045,liu2023transformation}. Following pre-training and fine-tuning, LLMs are primarily utilized by designing suitable prompts for various tasks. Leveraging powerful zero-shot capabilities, many tasks can be directly accomplished by guiding LLMs with straightforward prompts. For more complex tasks that cannot be achieved through simple prompts, a few-shot approach involving in-context learning is employed to guide LLMs in task completion. Additionally, incorporating chain-of-thought \cite{wei2022chain,kojima2022large} prompts in the prompt enhances in-context learning by introducing a reasoning process. The pipeline of the in-context learning and chain-of-thought is shown in Figure~\ref{fig:TheUltilizationofLLM}. In some specialized research directions, obtaining intermediate layer representations of LLMs may be necessary. For instance, in neuroscience studies, embedding representations from the model are used to investigate activation regions of brain functions \cite{qiang2023functional,he2023multi,liu2023spatial,oota-etal-2022-neural}.

Generally, there are several approaches to employing LLMs. The first involves accessing the capabilities of robust proprietary models through open API services, such as utilizing the API provided by ChatGPT \cite{openai2023gpt4}. The second approach includes deploying open-source LLMs for local use \cite{touvron2023llama}. The third method entails fine-tuning open-source LLMs to meet specific domain standards \cite{liu2023radiologygpt,liu2023radiologyllama2}, enabling their application in a particular field, and subsequently deploying them locally. In Table~\ref{tab:LLM}, we have compiled information on various open-source LLMs for reference. Researchers can choose from these open-source LLMs to deploy applications that best suit their needs.

\begin{table}[h]
	\caption{List of open source LLMs.}
    \label{tab:LLM}
	\centering
	\begin{tabular}{ccc}
	\toprule
	LLM & Size (B) & Links \\
	\midrule	
	T5 \cite{T5} & 11B & \url{https://github.com/google-research/text-to-text-transfer-transformer} \\
        CodeGen \cite{Codegen} & 16B & \url{https://github.com/salesforce/CodeGen}\\
        MOSS \cite{sun2023moss} & 16B  &  \url{https://github.com/OpenLMLab/MOSS}\\
        GLM \cite{zeng2022glm} & 130B & \url{https://github.com/THUDM/GLM}\\
        ChatGLM \cite{zeng2022glm} & 6B & \url{https://github.com/THUDM/ChatGLM3} \\
        ChatYuan \cite{clueai2023chatyuan} & 0.7B & \url{https://github.com/clue-ai/ChatYuan} \\
        OPT \cite{zhang2022opt} & 175B & \url{https://github.com/facebookresearch/metaseq} \\
        BLOOM \cite{workshop2022bloom} & 176B & \url{https://huggingface.co/bigscience/bloom} \\
        LLaMA \cite{touvron2023llama} & 65B & \url{https://github.com/facebookresearch/llama}\\
        CodeGeeX \cite{zheng2023codegeex} & 13B & \url{https://github.com/THUDM/CodeGeeX}\\
        Baichuan \cite{baichuan2023baichuan2} & 13B & \url{https://github.com/baichuan-inc/Baichuan2}\\
        Aquila & 7B &  \url{https://github.com/FlagAI-Open/FlagAI/tree/master/examples/Aquila}\\
        MiniGPT-4 \cite{zhu2023minigpt} & 25B & \url{https://github.com/Vision-CAIR/MiniGPT-4} \\
        Vicuna \cite{zheng2023judging} & 13B & \url{https://github.com/lm-sys/FastChat}\\
        \bottomrule
	\end{tabular}
\end{table}

\section{Future Directions and Implications}
\label{sec::future}
This section will delve into the future trends and impact of LLM technology. Our discussion will be structured into three parts: firstly, an exploration of the developmental trends within LLMs technology itself; secondly, an examination of the developmental directions for AI researchers; and finally, an analysis of the societal impact resulting from the ongoing development of LLMs. 

Based on existing experiences, it is evident that an ample supply of high-quality data and a sufficient number of parameters significantly contribute to enhancing the performance of models \cite{GPT3}. Looking ahead, the model scale of LLMs is expected to continue expanding, thereby augmenting their learning capabilities and overall performance. Moreover, the majority of currently available LLMs are confined to a single natural language modality, lacking extensions to process multimodal data such as images, videos, and speech. There is a potential future trajectory for LLMs to evolve towards handling information beyond text, incorporating multimodal data like images and audio. This evolution would empower models to comprehensively understand and generate multimodal content, significantly broadening the application scope of LLMs. The inevitable expansion of LLMs into the field of multimodality is bound to incur increased training costs. A pivotal focus for future developments lies in the efficient fine-tuning of parameters and the deployment of LLMs through techniques such as knowledge distillation, model compression, and quantization, aimed at reducing both the training and inference costs of LLMs. Another emerging trend is the domain-specific training and fine-tuning of LLMs for particular sectors, facilitating a more adept adaptation to and understanding of industry-specific terminologies and contexts. Lastly, in the exploration of potential new architectures for LLMs the current landscape predominantly relies on the transformer architecture. While the transformer architecture naturally boasts advantages such as parallel computing and adaptability to various input modalities, its design typically necessitates fixed-size inputs. This requirement may necessitate padding or truncation when dealing with variable-length sequences, potentially leading to computational and information inefficiencies, as well as challenges in generating coherent data. Investigating the potential of Recurrent Neural Network (RNN) architectures in the era of LLMs could emerge as a pivotal research direction. For instance, RWKV \cite{peng2023rwkv}, an LLM designed under the RNN architecture, has demonstrated competitive performance on various third-party evaluations, proving itself comparable to the majority of transformer-based LLMs. 

For researchers in the field of AI, working in isolation is becoming increasingly impractical. The future direction of AI development will intertwine with various industries, necessitating close collaboration with professionals from diverse fields. It is crucial to engage in collaborative efforts, bridging research disciplines, and collectively addressing challenges by combining expertise from different domains. Simultaneously, there is a fresh set of requirements for the comprehensive skills of AI researchers. Training and deploying LLMs necessitate proficiency in managing large-scale data and substantial practical experience in distributed parallel training. This criterion underscores the importance for researchers involved in LLM development to possess substantial engineering capabilities, addressing the challenges inherent in the process. Researchers who are interested in the field of LLMs must either possess engineering skills or adeptly collaborate with engineers to navigate the complexities of model development \cite{zhao2023survey}. 

As LLMs find widespread applications in societal life, concerns about ethical issues and societal impact are on a continuous rise. This may involve research and improvements in areas such as managing model biases and controlling the risk of misuse \cite{kaddour2023challenges}. Considering the paramount importance of privacy and data security, the future development of LLMs might involve more federated learning and decentralized approaches to enhance model performance while safeguarding user privacy. Developers should engage in interdisciplinary collaboration with experts from various fields, including decision-making, legal studies, and sociology, to establish standards and ethical frameworks for the development, deployment, and utilization of LLMs, mitigating potential harmful consequences. In terms of public awareness and education, mandatory awareness training should be implemented before large-scale public deployment and applications. This aims to enhance public understanding of the capabilities and limitations of LLMs, fostering responsible and informed use, especially in industries such as education and journalism.

\section{Conclusion}
\label{sec::conclusion}
The introduction of ChatGPT has ushered in a transformative era in the realm of Large LLMs, significantly influencing their utilization for diverse downstream tasks. The emphasis on cost-effective training and deployment has emerged as a crucial aspect in the evolution of LLMs. This paper has provided a comprehensive survey of the evolution of large language model training techniques and inference deployment technologies in alignment with the emerging trend of low-cost development. The progression from traditional statistical language models to neural language models, and subsequently to PLMs such as ELMo and transformer architecture, has set the stage for the dominance of LLMs. The scale and performance of these models, particularly exemplified by the GPT series, have reached unprecedented levels, showcasing the phenomenon of emergence and enabling versatile applications across various domains. Notably, the release of ChatGPT by OpenAI in November 2022 has marked a pivotal moment in the LLM landscape, revolutionizing the strength and effectiveness of AI algorithms. However, the current reliance on OpenAI's infrastructure underscores the necessity for alternative LLMs, emphasizing the need for domain-specific models and advancements in the training and deployment processes.

Training and deploying LLMs present challenges that demand expertise in handling large-scale data and distributed parallel training. The engineering capabilities required for LLM development highlight the collaborative efforts needed between researchers and engineers. As we explore the technical aspects of LLM training and inference in this review, it becomes evident that a deep understanding of these processes is essential for researchers venturing into the field. Looking ahead, the future of LLMs holds promising directions, including further advancements in model architectures, improved training efficiency, and broader applications across industries. The insights provided in this review aim to equip researchers with the knowledge and understanding necessary to navigate the complexities of LLM development, fostering innovation and progress in this dynamic field. As LLMs continue to evolve, their impact on natural language processing and AI as a whole is poised to shape the future landscape of intelligent systems.

\bibliographystyle{IEEEtran}
\bibliography{mybib}

\end{document}